\documentclass[letterpaper, 10pt, conference]{ieeeconf}  % Comment this line out if you need a4paper

\IEEEoverridecommandlockouts                              % This command is only needed if 

\usepackage[english]{babel}


\usepackage{amsmath, amssymb} % assumes amsmath package installed
\usepackage{accents}
\usepackage{cite}
 \usepackage{siunitx}
\usepackage{epsfig}
\usepackage{pst-all, graphics, graphicx, color}
\usepackage[crop=pdfcrop]{pstool}
\usepackage{psfrag}
\usepackage{subfigure}
\usepackage{epstopdf}
\usepackage{tikz}
\usepackage{multirow}
\usepackage{algorithm, algorithmic}
\usepackage{bm}
\usepackage{booktabs} % To thicken table lines
\usepackage{float}

\usepackage[nolist]{acronym}

\newcommand{\vb}[0]{\boldsymbol{b}}
\newcommand{\vc}[0]{\boldsymbol{c}}

\newcommand{\ve}[0]{\boldsymbol{e}}
\newcommand{\vf}[0]{\boldsymbol{f}}

\newcommand{\vp}[0]{\boldsymbol{p}}

\newcommand{\vx}[0]{\boldsymbol{x}}

\newcommand{\vz}[0]{\boldsymbol{z}}

\newcommand{\vA}[0]{\boldsymbol{A}}

\newcommand{\vC}[0]{\boldsymbol{C}}
\newcommand{\vD}[0]{\boldsymbol{D}}

\newcommand{\vF}[0]{\boldsymbol{F}}

\newcommand{\vH}[0]{\boldsymbol{H}}

\newcommand{\vK}[0]{\boldsymbol{K}}

\newcommand{\vM}[0]{\boldsymbol{M}}

\newcommand{\vP}[0]{\boldsymbol{P}}
\newcommand{\vQ}[0]{\boldsymbol{Q}}

\newcommand{\vW}[0]{\boldsymbol{W}}

\newcommand{\vGamma}[0]  {\boldsymbol{\Gamma}}

\newcommand{\vPhi}[0]    {\boldsymbol{\Phi}}

\newcommand{\tp}{{\mathsf{T}}}

\newacro{dmp}[DMP]{Dynamic Movement Primitive}
\newacro{seds}[SEDS]{Stable Estimator of Dynamical Systems}
\newacro{gmm}[GMM]{Gaussian Mixture Model}
\newacro{gmr}[GMR]{Gaussian Mixture Regression}
\newacro{gpr}[GPR]{Gaussian Process Regression}
\newacro{tpgmm}[TP-GMM]{Task-Parameterized \ac{gmm}}
\newacro{lfd}[LfD]{Learning from Demonstration}
\newacro{il}[IL]{Imitation Learning}
\newacro{wls}[WLS]{weighted least-squares}
\newacro{vs}[VS]{Visual Servoing}
\newacro{ilvs}[ILVS]{Imitation Learning based Visual Servoing}
\newacro{ds}[DS]{Dynamical System}
\newacro{rdsvs}[RDS-VS]{Reshaped \acl{ds} for \acl{vs}}
\newacro{rds}[RDS]{Reshaped \acl{ds}}
\newacro{clfdm}[CLF-DM]{Control Lyapunov Function-based Dynamic Movements}
\newacro{clf}[CLF]{Control Lyapunov Function}
\newacro{fdm}[FDM]{Fast Diffeomorphic Matching}
\newacro{vsds}[VSDS]{Variable Stiffness Dynamical System}

\definecolor{my_green}{rgb}{0.0,0.49,0.19}

\begin{document}

\title{\LARGE \bf
A Passivity-based Approach for Variable Stiffness Control with Dynamical Systems}

\author{Youssef Michel$^{1}$, Matteo Saveriano$^{2}$, and Dongheui Lee$^3$%
\thanks{$^{1}$Chair of Human-Centered Assistive Robotics, Technical University of Munich, Munich, Germany {\tt youssef.abdelwadoud@tum.de}}%
\thanks{$^{2}$Department of Industrial Engineering, University of Trento, Trento, Italy {\tt matteo.saveriano@unitn.it}.}%
\thanks{$^{3}$Autonomous Systems, TU Wien, Vienna, Austria and the Institute of Robotics and Mechatronics, German Aerospace Center (DLR), Munich, Germany {\tt dongheui.lee@tuwien.ac.at}}
}
%\markboth{IEEE Transactions on Robotics,~Vol.~x, No.~x, xxx~2015}%
%{Saveriano \MakeLowercase{\textit{et al.}}: Incremental Reshaping of Stable Dynamical Systems using Gaussian Process Regression}

\maketitle
%\thispagestyle{empty}
%\pagestyle{empty}

%%%%%%%%%%%%%%%%%%%%%%%%%%%%%%%%%%%%%%%%%%%%%%%%%%%%%%%%%%%%%%%%%%%%%%%%%%%%%%%%

\begin{abstract}

 % Variable Stiffness Dynamical Systems (VSDS) was proposed in our previous work \cite{} in order to follow the motion plan of a first order DS with a desired variable stiffness behavior, in a closed cloop configuration. 
 In this paper, we present a controller that combines motion generation and control in one loop, to endow robots with reactivity and safety. In particular, 
we propose a control approach that enables to follow the motion plan of a first order Dynamical System (DS) with a variable stiffness profile, in a closed loop configuration where the controller is always aware of the current robot state. This allows the robot to follow a desired path with an interactive behavior dictated by the desired stiffness. We also present two solutions to enable a robot to follow the desired velocity profile, in a manner similar to trajectory tracking controllers, while maintaining the closed-loop configuration. Additionally, we exploit the concept of energy tanks in order to guarantee the passivity during interactions with the environment, as well as the asymptotic stability in free motion, of our closed-loop system. The developed approach is evaluated extensively in simulation, as well as in real robot experiments, in terms of performance and safety both in free motion and during the execution of physical interaction tasks.  
 
 % The Variable Stiffness Dynamical Systems (VSDS) controller was proposed in our previous work in order to 
% In this letter, we propose a controller that enables to follow the motion plan of a first order Dynamical System (DS) with a variable stiffness profile, in a closed loop configuration \BL{where the controller is always aware of the current robot state}. The proposed controller allows to follow the 
% extends our Variable Stiffness Dynamical Systems (VSDS) approach \cite{chen2021closed} to further improve its safety and performance. First, we present two solutions to enable a robot to follow a desired velocity profile, in a manner similar to trajectory tracking controllers, while maintaining the closed-loop configuration. Second, we exploit the concept of energy tanks in order to guarantee the passivity during interactions with the environment, as well as the asymptotic stability in free motion, of our closed-loop system. The developed approach is evaluated extensively in simulation, as well as in real robot experiments, in terms of performance and safety both in free motion and during the execution of physical interaction tasks.  
 
 % In this work, we extend our controller capabilities . First, we present two solutions to enable the our robot to follow a desired velocity profile. Second, . Finally, our approach is extensively evaluated in simulations, as well on the real robot in terms of performance, task execution and safety ... 
\end{abstract}
\renewcommand{\abstractname}{Hol}

\begin{notetoprac}
The approach presented in this work allows for safe and reactive robot motions, as well as the capacity to shape the robot's physical behavior during interactions. This becomes crucial for performing contact tasks that might require adaptability or for interactions with humans as in shared control or collaborative tasks. Furthermore, the reactive properties of our controller make it adequate for robots that operate in proximity to humans or in dynamic environments where potential collisions are likely to happen.
\end{notetoprac}

% \begin{IEEEkeywords}
% x, xx, 
% \end{IEEEkeywords}

\IEEEpeerreviewmaketitle

\section{Introduction}\label{sec:intro}
Robots nowadays have witnessed a paradigm shift transitioning from the rigid and often position-controlled industrial manipulators to a safer, more compliant version, which enabled the seamless introduction of robots into domestic environments such as museums and hospitals where they co-exist in close proximity to humans~\cite{ajoudani2018progress}. This raises the need for planning and control frameworks capable of ensuring compliant and adaptive robot behaviors that ensure task execution while safely reacting to uncertainties. 

In this regard, Impedance Control  \cite{IC_HOG} is a prominent control approach capable of fulfilling such requirements. 
% In contrast to force or stiff position control, 
Impedance Control primarily aims at controlling the interactive robot behavior essentially regulating the energy exchange between the robot and its environment at the ports of interaction. A recent trend within the robotics community has focused on transitioning from constant to Variable Impedance Control (VIC) where the impedance parameters are allowed to vary over time/state~\cite{abu2020variable}. 
 This was inspired by human motor control theory which showed that humans continuously adapt their end-point impedance during physical interactions \cite{ContToolHap}, and further motivated by the increasing demand to explore more sophisticated interaction control techniques that would allow robots to adapt to different task contexts and environments. The authors in \cite{Rozo} and \cite{Peternel} use VIC in physical human robot collaboration for table carrying and cyclic sawing tasks, while in \cite{Lee11} it is exploited to realize incremental kinesthetic teaching. A framework was proposed in \cite{ABUDAKKA2018156} to learn variable stiffness profiles for carrying out contact tasks such as valve turning, and later extended in \cite{Michel} for bilateral teleoperation. Alternative approaches like~\cite{buchli2011variable, khader2020stability} use reinforcement learning to design variable stiffness profiles to optimize performance objectives such as energy and tracking errors.   
% This is realized by rendering a second order mass-spring damper system at the interaction point, often at the robot end-effector. 

 % For example, ... 

The requirement for safe and adaptive interaction controllers should be also complemented with motion generators that can guarantee flexibility and precision, while being reactive to possible perturbations or changes in the environment. Unfortunately, the classical way to command desired motions is to program the impedance controller and motion generator as two separate loops, where time-indexed trajectory generators such as splines or dynamic movement primitives~\cite{Saveriano2021Dynamic} feed the controller with a sequence of desired set-points parameterized with time. This is an open-loop configuration, where the motion generator does not have any feedback on the current robot state, and therefore clearly lacks robustness to temporal perturbations~\cite{kronander2015passive}. Furthermore, this raises major safety concerns in unstructured or populated environments where potential unplanned collisions would lead to very high forces or clamping situations that pose serious dangers to nearby humans, or lead to the damage of the robot or its environment~\cite{haddadin2009requirements}. 

To solve this problem, a potential solution is to combine motion generation and impedance control in the same control loop. The authors in \cite{Li99} suggested to encode a task via closed-loop velocity fields, and use a feedback control strategy based on virtual flywheels to ensure stability. In \cite{Khansari2014ModelingRD}, the authors were able to derive from the Gaussian Mixture Model formulation a closed loop impedance control policy with a state varying spring and damper, while deriving sufficient conditions for convergence. Along the same lines, the authors from \cite{Zadeh} proposed to shape the robot control as the gradient of a  non-parametric potential function learnt from user demonstrations. A different idea was presented in~\cite{ILOSA}, where the authors proposed an approach based on Gaussian processes to learn a combination of stiffness and attractors, in an interactive manner via  corrective inputs from the human teacher. 

Closely related to these works, first-order Dynamical Systems (DS) motion generators are becoming increasingly popular thanks to their stability features in terms of convergence to a global attractor regardless of the initial position. This is in addition to their flexibility in modeling a myriad of robotic tasks \cite{Saveriano13,catching,Amanhoud2019ADS}, and their ability to incorporate a wide range of machine learning algorithms, such as Gaussian mixtures models \cite{SEDS} and Gaussian Processes~\cite{kronander2015incremental}.   
Furthermore, the DS formulation naturally extends to closed-loop configuration control, allowing to fully exploit their inherent reactivity and stability properties. This was initially shown in~\cite{kronander2015passive}, where a passive controller was developed to track the velocity of a first order DS by selectively dissipating kinetic energy in directions perpendicular to the desired motion. However, the controller in~\cite{kronander2015passive} is flow-based and therefore does not possess the ability to restrict the robot along a desired path, nor the spring-like attraction behavior characteristic of stiffness. This was remedied in \cite{LAGS_22}, where the proposed control approach is still flow-based, however with the spring-like attraction behavior embedded inside the DS. 
% \textcolor{magenta}{@Youssef: Is the previous your VSDS work as well?}
In our previous work~\cite{chen2021closed}, we proposed our Variable Stiffness Dynamical Systems (VSDS) controller in order to embed variable stiffness behaviors into a first-order DS controlled in a closed-loop configuration. The controller requires the specification of a stiffness profile that can be provided by the user depending on the application, a first-order DS describing the motion plan, and consequently outputs a control force that follows the DS motion with an interactive behavior described by the stiffness, achieving also a spring-like attraction similar to \cite{LAGS_22}. Differently from \cite{LAGS_22}, however, the controller allows the complete decoupling of the stiffness profile specification from the desired motion, and can be also used with any first-order DS. Therefore, the approach combines the advantages of first-order DS in terms of flexibility and reactivity, with the interaction capabilities that can be achieved with variable stiffness behaviors. We showed the benefits of our controller in~\cite{chen2021closed} for autonomous task execution and in~\cite{Haotian} for shared control. \\
% In our previous works \cite{} and \cite{}, we showed the benefit of our \ac{vsds} controller for autonomous robot execution and in shared control, both in terms of safety and task success. 
% Unfortunately however, two drawbacks in the original \ac{vsds} formulation remained. 
In this work, we extend the original VSDS formulation with two important features. First, we aim to augment our controller with a trajectory tracking capability, such that the robot is able to follow both the position (as in the original VSDS) and the velocity profiles described by the first-order DS. The ability to accurately track trajectories on a dynamic level is one of the trademark characteristics of controllers relying on open-loop generated, time indexed trajectories, e.g., computed torque controllers~\cite{an1987experimental}. Incorporating such a feature in our VSDS provides means to accurately follow a motion profile while still benefiting from the inherit safety characteristic of the closed-loop configuration. \\
Our second goal is to propose a solution for guaranteeing the asymptotic stability (in free motion) and passivity (during physical interactions) of our VSDS controller. One of the main attributes that provide DS its appeal in modeling point-to-point motions is their ability to generate state trajectories with guaranteed asymptotic convergence to a target location, from anywhere in the state space. Therefore, it would be desirable that a controller designed to follow such a DS would also provide the same convergence proprieties. It is important to note that, since the controller operates in closed-loop, solely guaranteeing the asymptotic stability of the motion generation DS does not necessarily imply the overall stability of the robot motion. Instead, one has to additionally consider the controlled robot dynamics in the analysis. On the other hand, ensuring passivity is of paramount important for physically interacting robots, since it guarantees the stable interaction with arbitrary passive and possibly unknown environments \cite{Stefano}. To achieve these objectives, we exploit the concept of energy tanks commonly used to ensure passive behaviors, which were employed in a number of robotic applications including bilateral teleoperation~\cite{Franken}, hierarchical control~\cite{Dietrich2,MichelTRO}, and dynamical systems~\cite{kronander2015passive,saveriano2020energy}. \\
To summarize, with respect to our initial work in \cite{chen2021closed}, we present the following contributions:
\begin{itemize}
    \item We extend the original VSDS approach with a feedforward term that allows the tracking of a desired velocity profile. We propose and compare two different methods to achieve this objective. 
    \item We design a control strategy to ensure the asymptotic stability and passivity of our system. Therefore, we guarantee that the energy in the system remains bounded, and in consequence stability during physical interactions \cite{Stefano}, while also ensuring convergence to the equilibrium. To that end, we exploit 
    the use of energy tanks with a new formulation of the tank dynamics, and, additionally, a control term based on a novel form of a conservative potential field.
     \item We validate our developed approach extensively in simulations as well as on real robot experiments in a number of scenarios, where we also highlight the difference in performance compared to the original VSDS approach. % \cite{chen2021closed}. 
\end{itemize}
The rest of this work is organized as follows. Section~\ref{sec:background} provides background information and describes the VSDS formulation. In Sec.~\ref{sec:vsds_new}, we present our new formulation that permits velocity tracking. Passivity and stability of the proposed controller are analyzed in Sec.~\ref{sec:energy_control}. Experimental results are presented in Sec.~\ref{sec:experiments}, while section~\ref{sec:conclusion} states the conclusions and proposes further extensions.
\section{Preliminaries}\label{sec:background}
\textbf{Notation}: In the following, we use bold characters to indicate vectors and matrices. For a vector $\vp$, the notation $p^{i}$ indicates the $i$-th element of the vector, while for a matrix $\vP^{i,j}$ indicates the element at $i$-th row and $j$-th column. We use $\vP^{i,*}$ for $i$-th row and $\vP^{*,j}$ for the $j$-th column, and $\{\vp_k\}^{K}_{k=1}$ to indicate vector elements stacked together evaluated over $k=1...K$. Finally, we have $||\vp||=\sqrt{\vp^T\vp}$ indicate the $L_2$ norm, and $|\vp|$ for the element-wise absolute values. \\
The considered Cartesian-space gravity compensated dynamics of the robot can be expressed as
\begin{equation} \label{dynamics}
 \vM(\vx) \ddot{\vx} + \vC(\vx, \dot {\vx}) \dot{\vx} = \vF + \vF_{ext}
\end{equation}
where $\vx \in \mathbb{R}^m$ is the Cartesian state with $m$ as the number of DOF,  $\vM(\vx)$ is the Cartesian-space Inertia matrix, $\vC(\vx, \dot{\vx})$ is the Coriolis matrix, while $\vF$ and $\vF_{ext}$ correspond to the forces applied by controller and the external environment, respectively. \\
In our previous work \cite{chen2021closed}, we proposed our \ac{vsds} controller to compute $\vF$. The main idea behind the controller is to follow a path described by one of the  integral curves of a first order DS $\dot{\vx}_d=\vf_g(\vx)$, with an interactive behavior described by a desired stiffness profile $\vK_d(\vx)$. The first order DS $\vf_g(\vx)$ is assumed to be asymptotically stable around a global attractor $\vx_g$, and can be obtained e.g via learning with any state-of-art techniques such as Gaussian Mixture Models \cite{SEDS} or Gaussian Process Regression \cite{kronander2015incremental}.
The controller operates in closed-loop, in the sense that it is always aware of the current robot state, and also provides an attraction behavior around the reference path proportional to $\vK_d(\vx)$. The nominal motion plan represented by $\vf_g(\vx)$ represents a velocity field, and is assumed to have guaranteed asymptotic stability around a global attractor $\vx_g$. \\
In the original formulation, \ac{vsds} is designed as the nonlinear weighted sum of linear DS with dynamics $\vf_{l,i}(\vx)=\vA_i(\vx-\vx_{l,i})$ centered around a local attractor $\vx_{l,i}$ sampled from $\vf_g$, while $\vA_i$ is the stiffness of the $i$-th local computed via EigenValue Decomposition, as 
\begin{equation} \label{e.8}
    \vA_i = -\vQ_i \vK_{d,i} \vQ_i^\tp
\end{equation}
 where $\vK_{d,i}=\vK_{d}(\vx_{l,i})$ is a diagonal positive definite matrix, sampled from the desired stiffness profile. We design  $\vQ_i$ similar to \cite{kronander2015passive} by aligning the first eigenvector with $\frac{\vf_g(\vx)}{ ||\vf_g(\vx)|| }$, while the rest of the eigenvector are chosen perpendicular to the first one.  This means that $\vQ_i$ projects the stiffness values along and perpendicular to the current motion direction. \\
 We combine the linear DS via a Gaussian kernel for the $i$-th linear DS as
$
\omega_i(\vx) = {\rm exp}(-\frac{(\vx - \vx_{cen,i})^\tp (\vx- \vx_{cen,i})}{2(\epsilon^i)^2})
$
where $\vx_{cen,i} = \frac{1}{2}(\vx_{l,i} + \vx_{l,i-1})$ and $\epsilon^i$ as smoothing parameter proportional to the distance between the local attractors. The  weight of how each linear DS affects the dynamics is defined such that
\begin{equation} \label{omega}
    \widetilde{\omega}_i(\vx) = \frac{\omega_i(\vx)}{\sum^N_{j=1} \omega_j (\vx)}
\end{equation}
\ac{vsds} is then formulated as 
\begin{equation} \label{eq:VSDS_org}
    \vf_{vs,o}(\vx)= \sum^N_{i=1}\widetilde{\omega}_i(\vx) \vf_{i}(\vx). 
\end{equation} 
Finally, the control force sent to the robot is computed with 
\begin{equation} \label{F}
    \vF = \alpha(\vx)\vf_{vs,o}(\vx) - \vD(\vx)\dot{\vx}
\end{equation}
where $\alpha(\vx)$ is an optional state-dependent function that avoids large initial robot accelerations, while $\vD(\vx)\dot{\vx}$ is a dissipative field that can be simply designed with a constant positive definite damping matrix, or as we use in this work, a more elaborate design in order to assign a specific damping relationship to each linear DS i.e  
\begin{equation} \label{eq:D_i}
\vD(\vx)= \sum^N_{i=1}\widetilde{\omega}_i(\vx)\vD_i,
\end{equation}
where $\vD_i$ is a positive definite matrix. \\
% In our previous works \cite{} and \cite{}, we showed the benefit of our \ac{vsds} controller for autonomous robot execution and in shared control, both in terms of safety and task success. 
Unfortunately, the current VSDS formulation suffers from two main drawbacks. First, the robot only practically converges to the global attractor or very close to it, which is also partially dependent on parameter tuning. In other words, there is no theoretical guarantee that the robot is asymptotically stable with respect to the global attractor, which is one of the main features of first order DS. The second problem is that the velocity profile of the robot can arbitrary  differ from the velocity field represented by the original DS $\vf_g(\vx)$. Ideally, in an open-loop trajectory tracking problem, the velocity of the robot should be similar to that of the desired time-indexed trajectory $\vx_d(t)$, independently from the values of the stiffness and damping\footnote{Under the assumption of a stiffness high enough to overcome robot friction and properly tuned damping.}. These impedance parameters should on the other hand mainly affect the robot behavior in physical interaction, in the sense of how it reacts to perturbations or allows deviation from the desired trajectory.\\
In the next section, we will show our proposed approach to solve the two aforementioned problems, namely, tracking the velocity profile of $\vf_g(\vx)$, and ensuring the asymptotic stability/passivity of the closed loop system. Without loss of generality, we will assume in the following that the global attractor $\vx_g$ is shifted to the origin, such that the desired equilibrium of the system is at $(\vx=\mathbf{0},\dot{\vx}=\mathbf{0})$.
% \textcolor{magenta}{@Youssef: I would move this in the next section.}
% \YM{Done}
\section{Velocity Tracking VSDS}\label{sec:vsds_new}

\begin{figure}[!t]
\centering
\includegraphics[width =0.43\textwidth]{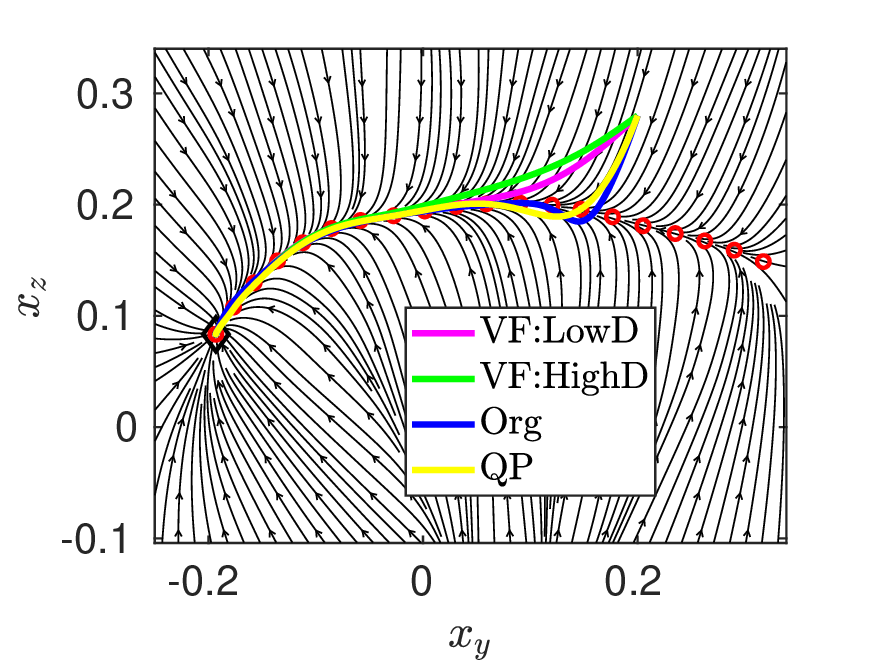}
\caption{Comparison between the symmetric attraction behavior of the original VSDS (Org), the optimization based approach (QP), the velocity feedback method with low (VF:LowD) and high (VF:HighD) damping gains.}
\vspace{-0.5\baselineskip}
\label{fig:symm_att}
\end{figure}
We start by proposing the following new formulation for our VSDS controller %\textcolor{red}{@Youssef: I'm thinking we probably do not need $\kappa(||\vx||)$ here!}
\begin{equation} \label{VSDS_new}
    \vf_{vs}(\vx) = \kappa(||\vx||)\left(\vf_{vs,o}(\vx) + \vf_{f}(\vx)\right) + \vPhi(\vx)
\end{equation}
\begin{equation} \label{F_new}
    \vF = \vf_{vs}(\vx) - \vD(\vx)\dot{\vx}
\end{equation}
where $\kappa(||\vx||)= 1 - \mathrm{e}^{-\alpha ||\vx||} $  is a smooth activation function that outputs a value of $0$ at the equilibrium, $\vPhi(\vx)$ is a conservative force field that also vanishes at the equilibrium, while $\vf_{f}(\vx)$ is another force field that ensures velocity tracking. The role and design of these controller elements will be further elaborated in the following.\\
As stated earlier, in a trajectory-tracking problem, it is desired that the robot not only follows the geometric path described by the trajectory, but also follows its timing law, which is mainly highlighted by the velocity of the robot being close or identical to that of the desired trajectory. This should also happen independently of the chosen stiffness profile. To achieve this objective and follow the velocity of the trajectory described by $\vf_g(\vx)$, we design the force field $\vf_f(\vx)$ from eqn. \eqref{VSDS_new} accordingly. This force field can be viewed as a feed forward term, similar to those typically used in computed torque control methods~\cite{an1987experimental}, that rely on the desired trajectory higher-order derivatives to ensure trajectory tracking. In the following, we propose two different solutions for the design of $\vf_f(\vx)$ to achieve this desiderata. 
\subsection{Velocity Feedback Approach}
The first possibility we explored in this regard was to simply augment our formulation with a velocity tracking term. The solution is inspired by the passivity-based controller proposed in~\cite{kronander2015passive} that follows the integral curves of a first order DS with a velocity feedback action. The control law can be designed with the feed-forward term as: 
\begin{equation} \label{eq:VSDS_tracking}
    \vf_f(\vx)=\vD(\vx)\vf_g(\vx)
\end{equation}
with $\vD(\vx)$ designed according to \eqref{eq:D_i}, and 
where, together with \eqref{F_new}, we are closing the loop around the velocity tracking error $\dot{\ve}=\vf_g(\vx)-\dot{\vx}$. \\
While this solution is simple and straightforward to implement, we noticed that whereas high damping gains lead to better velocity tracking, they result in slightly altering the symmetric attraction behavior compared to the original VSDS, which indicates the robot ability to attract back to the path when perturbed. Furthermore, as it will be explained in the experimental validation of section V.B, the formulation also results in higher contact forces during external interactions. 
% \begin{figure}[!t]
% \centering
% %\includegraphics[width =0.4\textwidth]{pics/stiffness.png}
%  \subfigure{
%         \includegraphics[width =0.225\textwidth]{img/DiffStiffProf_Pos.eps}
%     }
%     \subfigure{
%         \includegraphics[width =0.225\textwidth]{img/DiffStiffProf_Vel.eps}
%     }
% \caption{Simulation Results of a Mass Spring Damper for different stiffness profiles} 
% \vspace{-0.5\baselineskip}
% \label{stiffness}
% \end{figure}

\subsection{Optimization-based design}
To alleviate this problem, the second solution we propose is to optimize the feed-forward force fields based on some optimal reference behavior. More specifically, our aim is that a robot driven by a VSDS control law is to follow a reference path, as well as a desired velocity, in a manner similar to time-indexed open-loop trajectory tracking. In other words, in free motion, the simplest form of the optimal target behavior is equivalent to: 
\begin{equation} \label{desired_dynamics}
 \vM(\vx) \ddot{\vx} =\vK_d(\vx)(\vx_d(t)-\vx)-\vD_d(\vx)\dot{\vx},
\end{equation}
where $\vD_d(\vx)$ is a damping profile computed from the stiffness profile in order to maintain a critically damped system (or to achieve another control objective), while $\vx_d(t)$ is a time-indexed trajectory generated by the open-loop integration of $\vf_g(\vx_d)$ starting from the robot initial position. In principle, we could have included also $\dot{\vx}_d(t)$ in (\ref{desired_dynamics}); however, we found that (\ref{desired_dynamics}) in its current form results in good tracking results.  \\
We aim to design $\vf_f(\vx)$ based on \eqref{desired_dynamics}. Assuming that $\vf_g(\vx)$ and $\vK_d(\vx)$ are available, the second-order system \eqref{desired_dynamics} can be simulated offline. This results in a data-set that consists of $\{\vx_t\}_{t=0}^{T}$ and $\{\vF_{s,t}\}_{t=0}^{T}$, with $t$ as the time index and $T$ is the total simulation time. The simulated position response of the second-order system is $\vx_t$, while $\vF_{s,t}=\vK_d(\vx_t)(\vx_{d,t}-\vx_t)$ is the resulting spring force. We can then optimize $\vf_f(\vx)$ offline based on the collected data. We propose the following structure for $\vf_f(\vx)$: 
\begin{equation} \label{eq:ff_qp}
    \vf_f(\vx)=\sum^N_{i=1}\widetilde{\omega}_i(\vx) \vGamma_i. 
\end{equation}
which represents a weighted sum of constant forces $\vGamma_i$. The goal of the optimization can then be formulated as
\begin{subequations} \label{eq:opt1}
\begin{alignat}{2}
&\!\min_{\vGamma_{i,\,i=1,\ldots,N}}   &\quad& ||\vf_{vs}(\vx_t)-\vF_{s,t}||^{2}\label{eq:optProb}\\
&\text{subject to} & & \underline{\vGamma}_{i} \leq \vGamma_i \leq \overline{\vGamma}_{i},\label{eq:constraint1}\\
&  &      &
|\vf_{vs}(\vx_0)|>\underline{\vF} ,\label{eq:constraint2}
\end{alignat}
\end{subequations}
% \MS{(12c) does $|\cdot|$ indicate the element-wise absolute value? Or the norm? In the latter case we should use $\underline{F}$.} \YM{Actually it is the element-wise absolute values}
which minimizes the norm between the VSDS term $\vf_{vs}(\vx_t)$ over the simulated position response and the resulting spring term $\vF_{s,t}$ of the second order system. The constraint \eqref{eq:constraint1} was added to ensure reasonable upper and lower bounds on the constant force terms. On the other hand, the constraint \eqref{eq:constraint2} ensures a high enough initial spring force that can overcome robot friction, which becomes crucial for the implementation of the control policy on the real robot. \\
To solve \eqref{eq:opt1}, tools such as \texttt{fmincon} provided by Matlab\textsuperscript{\tiny\textregistered} can be used. This however resulted in a high computation time (around 1 minute) in order to compute the optimal solution, which also increases as the number of linear DS $N$ in the VSDS increases. To improve efficiency, in the following we show that it is possible to formulate our optimization as a convex Quadratic Program (QP). First, let us write the problem \eqref{eq:opt1} as 
\begin{subequations} \label{eq:opt2}
\begin{alignat}{2}
&\!\min_{\vGamma_{i,\,i=1,\ldots,N}}   &\quad& \Vert\sum^N_{i=1}\kappa(\Vert\vx_t\Vert)\widetilde{\omega}_i(\vx_t) \vGamma_i+\vf_s(x_t)-\vF_{s,t}\Vert^{2}\label{eq:cost2}\\
&\text{subject to} & & \underline{\vGamma}_{i} \leq \vGamma_i \leq \overline{\vGamma}_{i},\label{eq:constraint12}\\
&  &      &
|\sum^N_{i=1}\kappa(||\vx_0||)\widetilde{\omega}_i(\vx_0)\vGamma_i+\vf_{s}(\vx_0)|>\underline{\vF} ,\label{eq:constraint22}
\end{alignat}
\end{subequations}
where  $\vf_s(\vx)=\vf_{vs,o}(\vx)+\vPhi(\vx)$. Defining :
\begin{subequations}
\begin{alignat}{2}
&\vW(\vx)=  \begin{bmatrix}
\kappa(||\vx_0||)\widetilde{\omega}_1(\vx_0)  & \dots  & \kappa(||\vx_0||)\widetilde{\omega}_N(\vx_0)  \\
\vdots & \ddots & \vdots\\
\kappa(||\vx_T||)\widetilde{\omega}_1(\vx_T)  & \dots  & \kappa(||\vx_T||)\widetilde{\omega}_N(\vx_T),
    \end{bmatrix}\\
    &\vW_a=\begin{bmatrix} 
    \vW(\vx) & \mathbf{0}  \\
         \mathbf{0} & \vW(\vx)
    \end{bmatrix} ,    \vW_{a,0}=\begin{bmatrix} 
    \sigma_1\vW^{1,*} & \mathbf{0} \\
   \mathbf{0}  &   \sigma_2\vW^{1,*} \\
    \end{bmatrix},   \\
%    &\vf_s(\vx)=\vf_{vs,o}(\vx)+\vPhi(\vx)\\
   & \underline{\vGamma}_{sh}=\begin{bmatrix}
        \underline{F}^1 -\sigma_1f^1_s(\vx_0)\\
        \underline{F}^2 -\sigma_2f^2_s(\vx_0)\\
    \end{bmatrix} 
    \\
   & \vF_{sh,t}=\vF_{s,t} -\vf_s(\vx_t)\\
    &\vF_{sh}=\begin{bmatrix} \{ F_{sh,t}^{1} \}^{T}_{t=1}  & \{ F_{sh,t}^{2} \}^{T}_{t=1} \end{bmatrix} 
\end{alignat} 
\end{subequations}
Generally, we have that $\vW \in \mathbb{R}^{T\times N}$, $\vW_a \in \mathbb{R}^{mT \times mN}$, 
$\vW_{a,0} \in \mathbb{R}^{m\times mN}$, $\underline{\vGamma}_{sh} \in \mathbb{R}^{m }$ and $\vF_{sh} \in \mathbb{R}^{mT}$. For ease of illustration, we chose to formulate all the above terms with $m=2$.
The binary variable $\sigma_i, i=1,2$ is defined as 
\begin{equation}
\sigma_i=
\begin{cases}
	1   \quad f^i_s(\vx_0) \geq 0 \\ 
	-1  \quad f^i_s(\vx_0)<0 
	\end{cases}.
 \label{eq:sigma}
\end{equation}
we can then re-write \eqref{eq:opt2} as 
\begin{subequations} \label{eq:opt3}
\begin{alignat}{2}
&\!\min_{\vz}   &\quad& (\vW_a\vz - \vF_{sh,t})^\tp(\vW_a\vz - \vF_{sh,t})     \label{eq:optProb3}\\
&\text{subject to} & & \underline{\vz} \leq \vz \leq \overline{\vz},\label{eq:constraint31}\\
&  &      &
\vW_{a,0}\vz>\underline{\vGamma}_{sh} ,\label{eq:constraint32}
\end{alignat}
\end{subequations}
with $\vz=\begin{bmatrix} \{ \Gamma^{1} \}^{N}_{i=1}  & \{ \Gamma^{2} \}^{N}_{i=1} \end{bmatrix}$,
$\underline{\vz}=\begin{bmatrix} \{ \underline{\Gamma}_i^{1} \}^{N}_{i=1}  & \{ \underline{\Gamma}_i^{2} \}^{N}_{i=1} \end{bmatrix}$  and $\overline{\vz}=\begin{bmatrix} \{ \overline{\Gamma}_i^{1} \}^{N}_{i=1}  & \{ \overline{\Gamma}_i^{2} \}^{N}_{i=1} \end{bmatrix}$ containing the elements of all DOF concatenated. The introduction of the binary variable \eqref{eq:sigma} is necessary to realize the constraint 
\eqref{eq:constraint22} which can be reformulated as $\sum^N_{i=1}\kappa(||\vx_0||)\widetilde{\omega}_i(\vx_0)\vGamma_i>\underline{\vF}-\vf_s(\vx_0)$ if $\vf_s(\vx_0)\geq 0$,
and $\sum^N_{i=1}-\kappa(||\vx_0||)\widetilde{\omega}_i(\vx_0)\vGamma_i>\underline{\vF}+\vf_s(\vx_0)$ otherwise.  Since both constraints cannot be active at the same time, the role of $\sigma_i$ is to activate one of these constraints depending on the sign of the initial force $\vf_s(\vx_0)$, resulting in the constraint formulated in \eqref{eq:constraint32}. \\
Expanding further, and through some simple manipulations, we can reformulate \eqref{eq:opt3} as 
\begin{subequations} \label{eq:QP}
\begin{alignat}{2}
&\!\min_{\vz}   &\quad& \frac{1}{2}\vz^\tp\vH\vz + \vc^\tp\vz\label{eq:costqp}\\
&\text{subject to} & & \vA\vz \geq \vb,\label{eq:constraint1qp}\\
&  &      &
\underline{\vz} \leq \vz \leq \overline{\vz} .\label{eq:constraint2qp}
\end{alignat} 
\end{subequations}
 where $\vH=\vW_a^\tp\vW_a$, $\vc=-\vW_a^\tp \vF_{sh}$,
$\vA=\vW_{a,0}$, $\vb=\underline{\vGamma}_{sh}$, which represents the well-known form of a QP program. It is worth mentioning  that, when expanding \eqref{eq:optProb3}, a constant term that does not depend on $\vz$ has been omitted, since it does not affect the optimization.
 Finally, it is worth noting that the reformulation of the optimization as a QP-program results in a substantial reduction of the computation time for the optimization, which now gets solved in approximately $0.1\,$s, for $N=20$. \\
Figure \ref{fig:symm_att} shows the resulting attraction behavior of a simulated mass slightly perturbed from the reference path, and driven by a control law designed based on \textit{i)} the original VSDS formulation (eq.~\eqref{eq:VSDS_org}), \textit{ii)} the QP-based optimization (eq.~\eqref{eq:ff_qp}) and Velocity feedback (eq.~\eqref{eq:VSDS_tracking}) methods for the design of $\vf_{f}(\vx)$, 
with the eigenvalues of $\vD(\vx)$ where set once to \textit{iii)} low and once to \textit{iv)} high. While the behavior with the original approach and the QP methods are similar, for the velocity feedback method, increasing the damping gain delays the contact point between the mass and the reference path. This can be attributed to $\vf_g(\vx)$ which naturally points towards the global attractor. Increasing $\vD(\vx)$ magnifies this behavior and neutralizes the effect of $\vf_{vs,o}(\vx)$ which aims to pull the mass back to the reference path in a spring-like manner, and therefore interferes to some extent with the original VSDS dynamics.

\section{Energy Tank-based Control}\label{sec:energy_control}
The \ac{vsds} formulation developed in Sec.~\ref{sec:vsds_new} is exploited to drive the robot in closed-loop, i.e., we need to explicitly take into account the robot dynamics to derive stability results. In general, the closed-loop implementation of \ac{vsds} is not guaranteed to drive the robot towards a desired equilibrium neither to ensure a safe interaction with a passive environment. In this section, we analyze the stability and passivity properties of \ac{vsds} and exploit the energy tanks formalism to ensure \textit{i)} asymptotic stability in free motion and \textit{ii)} a passive interaction behavior. 
% \textcolor{red}{@Matteo: I think we should mention that the difference to your approach (and maybe others) is that we consider the robot dynamics as well }\MS{Check now...}

\begin{figure}[!t]
\centering
	
	\includegraphics[width=0.48\columnwidth]{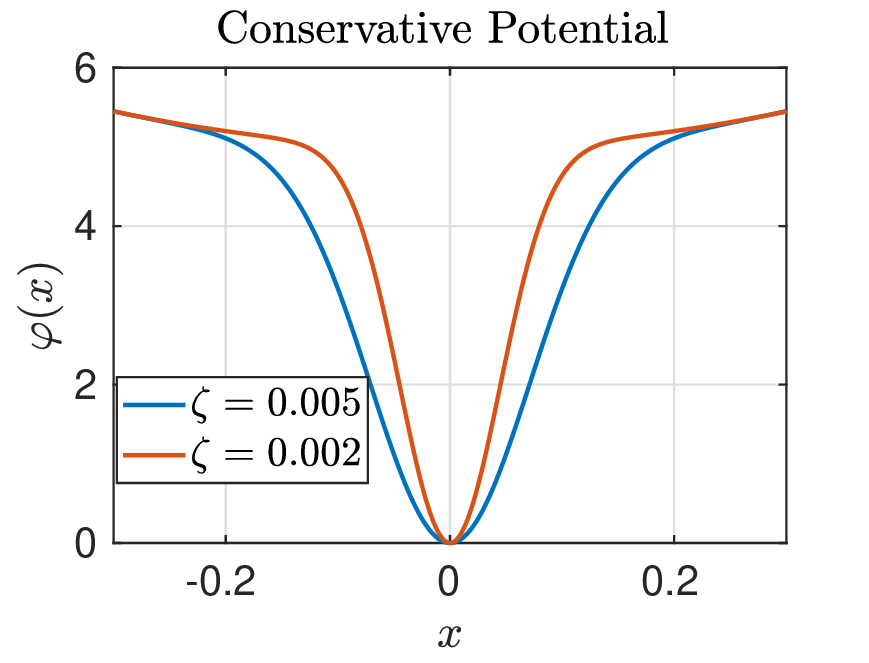}
		\label{fig:Potential_field}
	\includegraphics[width=0.48\columnwidth]{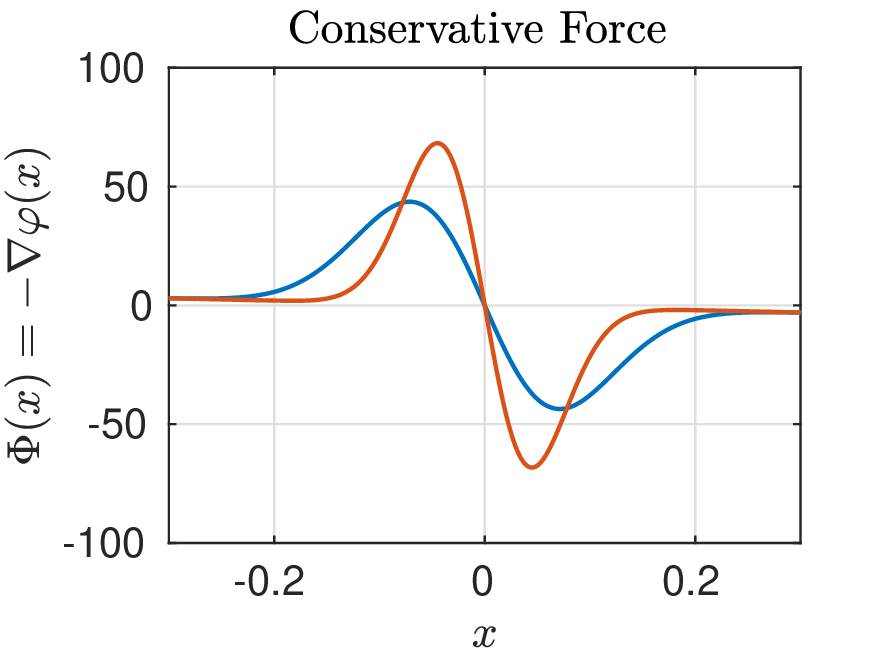}
		\label{fig:Potential_force}
	
 \caption{Left: Illustration of the conservative potential over 1 DOF for two values of $\zeta$. Right: The corresponding force field derived as the gradient of the potential. In both cases, we use $\kappa_o=5$ and $\tau_{min}=5$}.
	\label{fig:Potential}
\end{figure}

\subsection{Passivity analysis}\label{subsec:passivity}
We analyze the passivity of the system under the \ac{vsds} control law \eqref{VSDS_new}.
%In this work, the passivity of a system is defined as
%\begin{definition}
%xxxx
%\end{definition}
The closed-loop dynamics obtained by substituting~\eqref{F_new} and~\eqref{VSDS_new} into the robot dynamics~\eqref{dynamics} is %(we omit the state dependency for simplicity)
\begin{equation}
    \vM(\vx)\ddot{\vx} = -\vC(\vx,\dot{\vx})\dot{\vx} + \vF_{ext} + \vf_{vs} - \vD(\vx)\dot{\vx},
    \label{closed_loop_dyn}
\end{equation}
% where we defined the dissipative field in~\eqref{F} as $\vPsi(\vx,\dot{\vx}) = \sum^N_{i=1}\widetilde{\omega}_i(\vx)\vD_i\dot{\vx} = \vD(\vx)\dot{\vx}$.
where $\vD(\vx)$ is positive definite if each $\vD_i$ is a positive definite matrix.\\
We consider the storage function
\begin{equation}
    \mathcal{W}(\vx,\dot{\vx}) = \frac{1}{2}\dot{\vx}^\tp \vM(\vx)\dot{\vx} + \varphi(\vx),
    \label{storage_function}
\end{equation}
where the first term on the right site is the kinetic energy and $\varphi(\vx)$ is the potential function that generates the conservative field $\vPhi(\vx)$, i.e., $\vPhi(\vx) = -\nabla\varphi(\vx)$, and where we have $\varphi(\bm{0})=0$ and $\varphi(\vx)>0$ $\forall \vx \neq \bm{0}$. 
%\textcolor{blue}{Added some details regarding the potential} \MS{I would move this in the contributions right after eq (7)-(8) mentioning it is used for stability reasons.} \YM{equations (7-8) are now in the velocity tracking section, I feel adding details regarding the potential would be a deviation from the main message of that section  }
We design $\varphi(\vx)$ as
\begin{equation}
    \label{eq:constpot}
    \varphi(\vx)=k_o(1-e^{-\frac{\vx^\tp\vx}{2\zeta}}) +\tau_{min}\vx^\tp\vx
\end{equation}
%\YM{@Matteo is this formulation correct ?}
where $\tau_{min}$, $k_o$ and $\zeta$ are positive constants. In principle, we could have used a simple potential with a constant spring $\vK_c$ such that $\varphi=\frac{1}{2}\vx^\tp\vK_c\vx$. This design however would interfere with the VSDS dynamics and the desired stiffness behavior specified by $\vK_d$. On the other hand, the choice in \eqref{eq:constpot} allows us to selectively tune the effect of the conservative potential. For example, we can choose to have a weak influence for the potential in regions far away from the equilibrium by choosing a low $\tau_{min}$, while smoothly transitioning to have stronger influence in a small neighborhood around the attractor to ensure convergence. The width and strength of this neighborhood are controlled by the parameters $\zeta$ and $k_o$, as depicted in Fig. \ref{fig:Potential}.    

Taking the time derivative of $\mathcal{W}$ and considering the expression of $\vM(\vx)\ddot{\vx}$ in~\eqref{closed_loop_dyn} we obtain (omitting the arguments for simplicity)
\begin{equation}
	\begin{split}
    \dot{\mathcal{W}} &= \dot{\vx}^\tp \vM\ddot{\vx} + \nabla\varphi^\tp\dot{\vx} \\
    									&= -\dot{\vx}^\tp \vD(\vx) \dot{\vx} + \dot{\vx}^\tp \vF_{ext} + \kappa\dot{\vx}^\tp\left(\vf_{vs,o} + \vf_{f}\right)
    \end{split},
    \label{storage_function_derivative}
\end{equation}
where we used the property that $\dot{\vM} - 2\vC$ is skew-symmetric. The sign of $\kappa\dot{\vx}^\tp\left(\vf_{vs,o} + \vf_{f}\right)$ is undefined and does not allow to conclude the passivity of the system.

% \begin{figure}[!t]
% \centering
% 	\subfigure []{
% 	\includegraphics[width=0.22\textwidth]{img/Error_OrgVsNew.eps}
% 		\label{fig:coll_pos}
% 	}
% 	\subfigure []{
% 	\includegraphics[width=0.22\textwidth]{img/E_tank_new.eps}
% 		\label{fig:coll_force}
% 	}
%  \caption{Simulation Results old vs new \textcolor{red}{@Matteo: For now to validate the results, we can depending on space add an entire simulation section}}
% 	\label{fig:collisions}
% \end{figure}

\subsection{Energy tank-based passification}\label{subsec:passification}
We resort to the concept of energy tanks ~\cite{kronander2015passive,secchi2006position, lee2010passive} to render to closed-loop dynamics~\eqref{closed_loop_dyn} passive. The idea of energy tanks is to recover the energy dissipated by the system and use it to execute locally non-passive actions without violating the overall passivity. To this end, we consider an energy storing element with storage function $s(\vx,\dot{\vx})$. The dynamics of $s$ is defined as
\begin{equation}
\dot{s} = \alpha(s) \dot{\vx}^\tp \vD(\vx)\dot{\vx} - \beta(z,s)z - (\eta - \kappa(\Vert \vx \Vert))s \label{s_dot},
\end{equation}
where $z=\kappa(\Vert \vx \Vert)\dot{\vx}^\tp\left(\vf_{vs,o} + \vf_{f}\right)$ is the term with undefined sign in~\eqref{storage_function_derivative}. The novelty in our formulation lies in the term $- (\eta - \kappa(\Vert \vx \Vert))s$ with $\eta > 1$ which ensures that $s \rightarrow 0$ for $\dot{\vx} = 0$. This property is exploited  in Sec.~\ref{subsec:stability} to show the stability of the closed-loop system. Assuming that  $\eta \approx 1$ also reduces the effects of $- (\eta - \kappa(\Vert \vx \Vert))s$ on tank dynamics far from the position equilibrium (e.g., $\vx = \bm{0}$) while ensuring a rapid convergence of $s$ approaching the desired state ($\kappa(0) = 0$). %The tank dynamics in~\eqref{s_dot} forces the energy $s$ to vanish at the equilibrium, i.e., $s(\bm{0},\bm{0})=0$. More in detail, close to the equilibrium the dynamics of $s$ is governed by $\dot{s}_1$ that depletes the energy.
\begin{figure*}[t]
\centering
	\subfigure []{
	\includegraphics[width=0.25\textwidth]{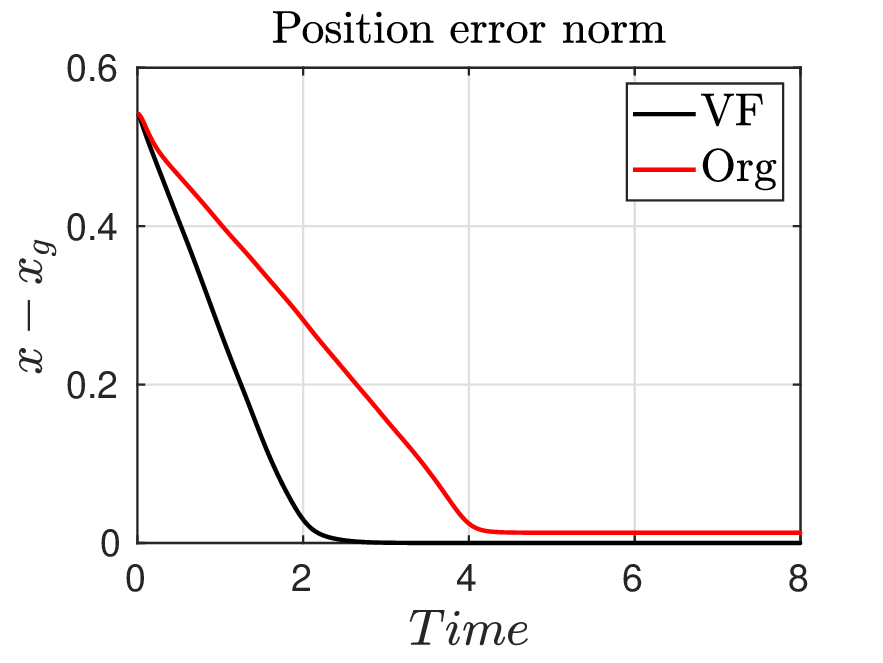}
		\label{fig:sim_norm_curve}
	}
	\subfigure []{
	\includegraphics[width=0.25\textwidth]{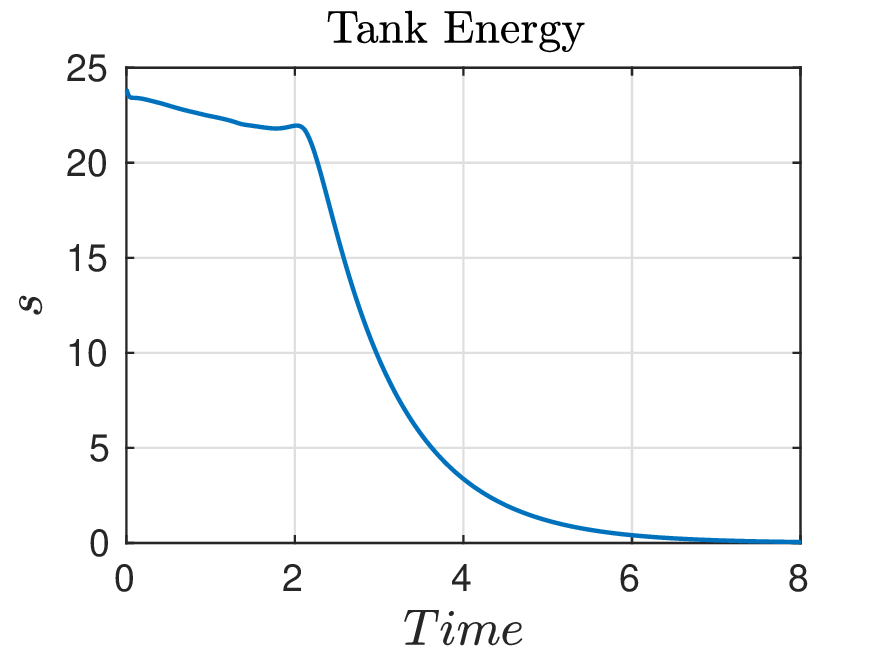}
		\label{fig:tank_curve}
	}
 \subfigure []{
	\includegraphics[width=0.25\textwidth]{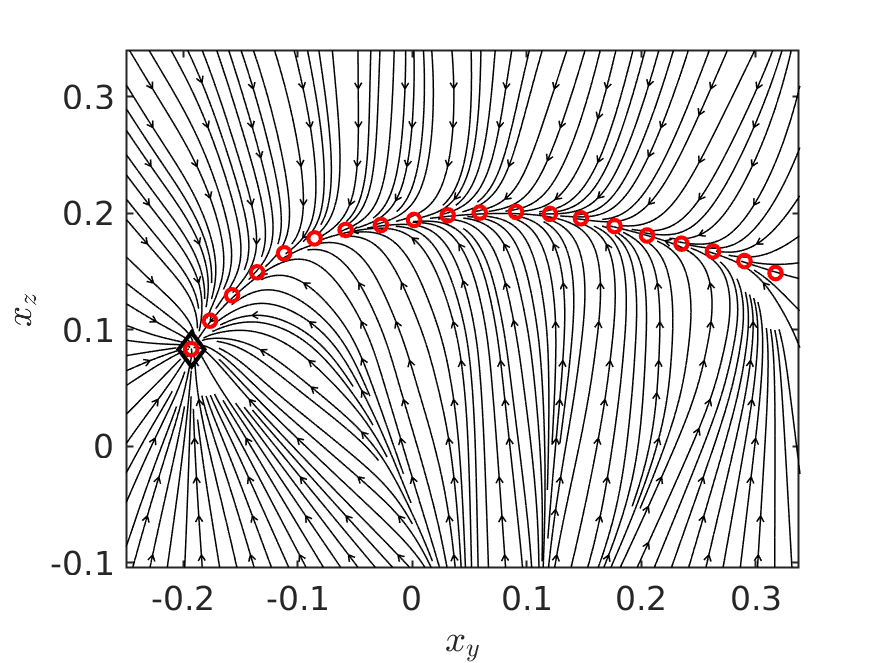}
		\label{fig:curve_stream}
	}
 \\
  \subfigure []{
 \includegraphics[width=0.25\textwidth]{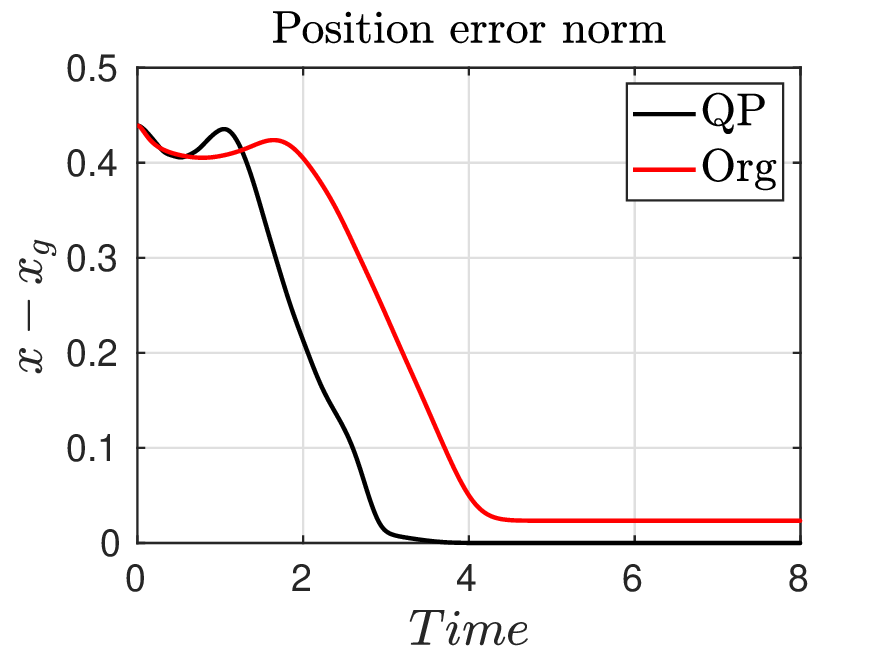}
		\label{fig:angle_norm}
	}
  \subfigure []{
 \includegraphics[width=0.25\textwidth]{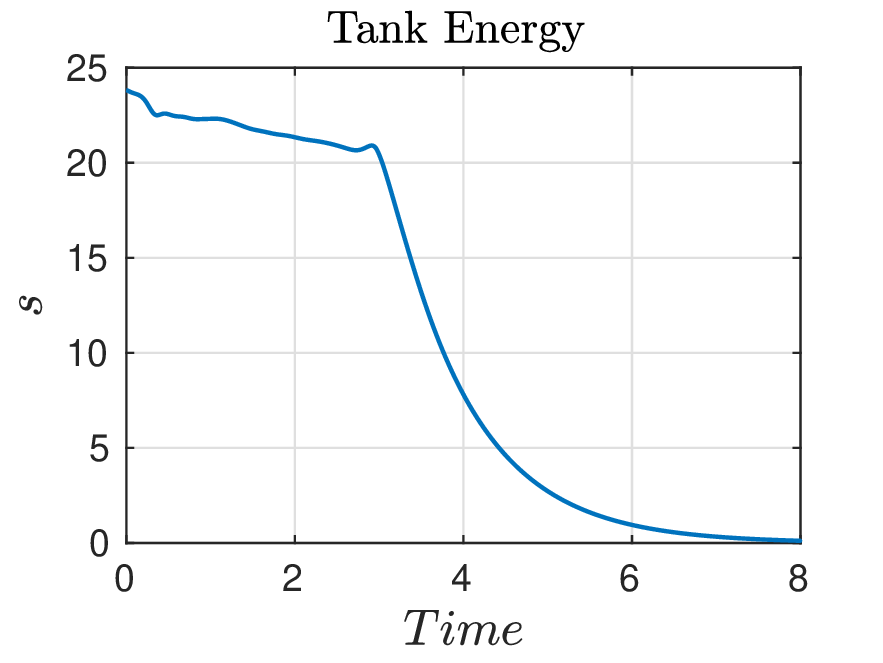}
		\label{fig:angle_tank}
	}
  \subfigure []{
  \includegraphics[width=0.25\textwidth]{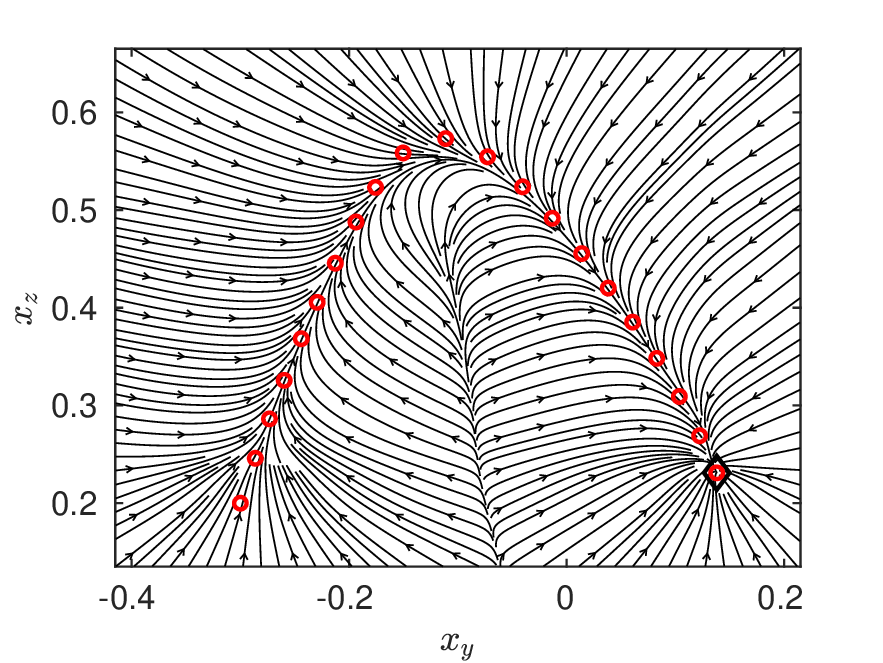}
		\label{fig:angle_stream}
	}  
 \\
  \subfigure []{
  \includegraphics[width=0.25\textwidth]{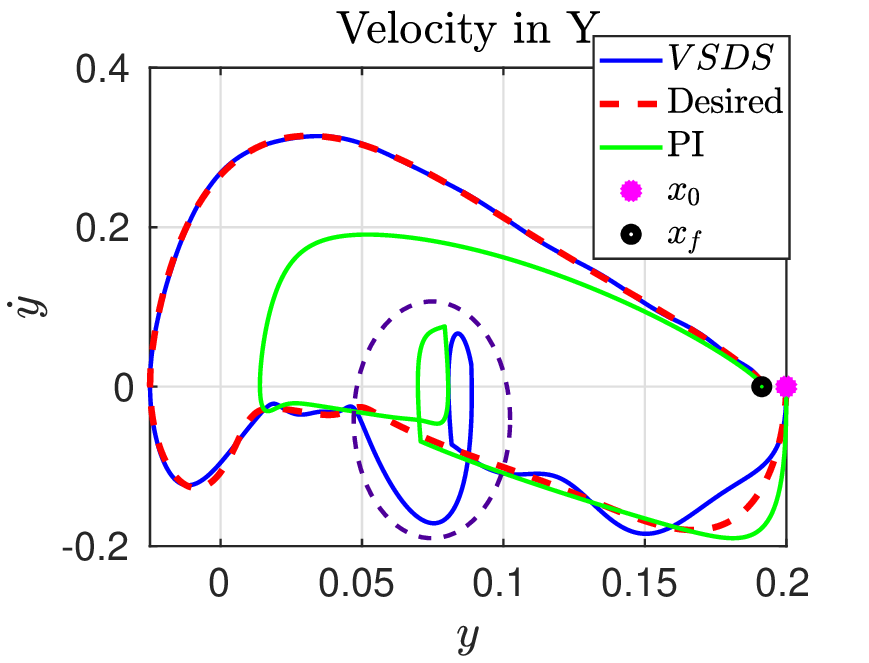}
		\label{fig:w_vel_y_comp}
	} 
  \subfigure []{
 \includegraphics[width=0.25\textwidth]{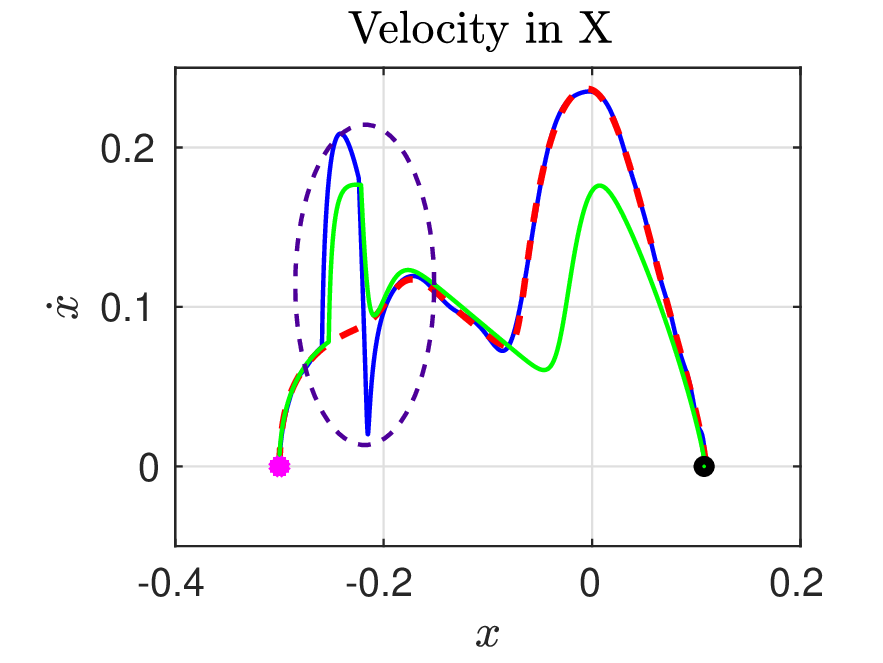}
		\label{fig:w_vel_x_comp}
	}
   \subfigure []{
 \includegraphics[width=0.25\textwidth]{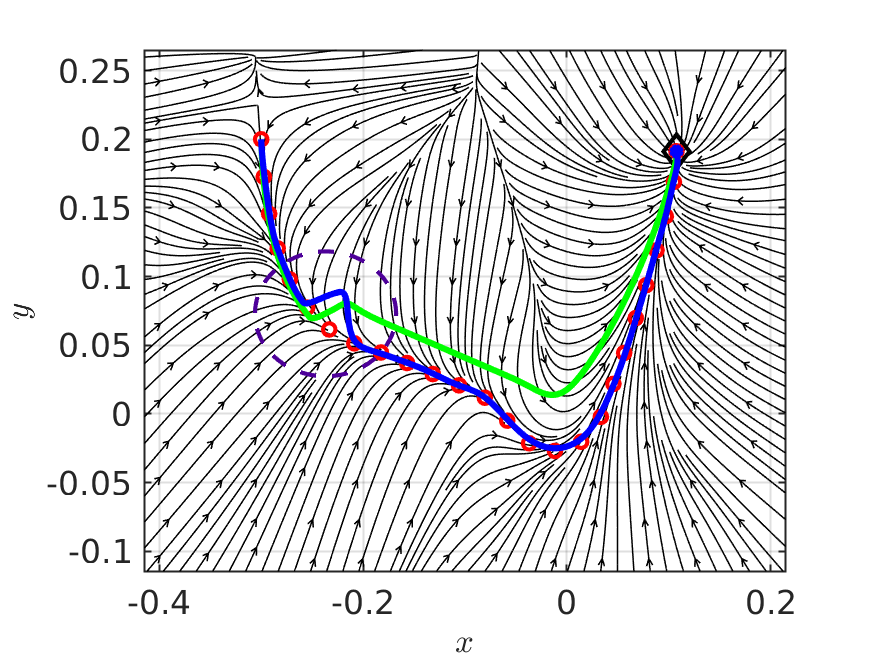}
		\label{fig:w_spatial_comp}
	}
 	\caption{Simulation results for the VSDS controller applied on a simulated mass. The first row shows the comparison results of the original VSDS controller (Org in the legend), to the velocity feedback formulation for computing the feed forward terms (VF), applied on a curve motion. The second row shows the comparison between the orignal formulation and the QP-optimization approach (QP), applied on an angle shaped motion. The streamlines of the VSDS dynamics are shown in the last column, with red dots depicting the local attractors, and the rhombus the global equilbruim. The last row shows the comparison results between the QP controller (blue), and the controller from \cite{kronander2015passive} (green). The plots (g) and (h) show state space plots of the velocity vs position in the $y$ and $x$ directions, and where the red dotted line depicts the reference state trajectory, the pink circle is the initial state, while the black dot is the final state.    }
	\label{fig:simres}
\end{figure*}
The variables $\alpha(s)$ and $\beta(z,s)$ satisfy
\begin{equation}
	\begin{cases}
		0 \leq \alpha(s) < 1 \quad s < \overline{s} \\ 
		\alpha(s) = 0 ~~~\qquad \text{otherwise}
	\end{cases},
	\label{eq:conditions_alpha}
\end{equation}
%where $\alpha(s)$ is strictly smaller than $1$ to ensure global stability (see Sec.~\ref{subsec:stability}). The scalar gain $\beta(z,s)$ satisfies
and
\begin{equation}
	\begin{cases}
    	\beta(z,s) = 0 ~~\qquad s \geq \overline{s} ~\text{and}~ z < 0 \\ 
		\beta(z,s) = 0 ~~\qquad s \leq 0~\text{and}~ z \geq 0 \\ 
		0 \leq \beta(z,s) \leq 1 \quad \text{otherwise}
	\end{cases},
	\label{beta}
\end{equation}
where $\overline{s}$ is the maximum allowed energy %\YM{is it $\overline{s}=\kappa s_0$ ?}\MS{I think it is not needed in the new formulation, can be also another value}. 
This definition of $\alpha(s)$ and $\beta(z,s)$ ensures that $s \geq 0$ everywhere if the initial energy $s_0 \geq 0$.  
%
%The tank dynamics in~\eqref{s_dot} forces the energy to be $s \geq 0$ everywhere. 
Therefore, we can add $s$ to the storage function in~\eqref{storage_function} as
% \YM{Should we mention here that we consider now the new equilbr. to be $(\vx, \dot{\vx},s) = (\bm{0},\bm{0},0)$ ?  this also verifies that the lyaponov vanishes at (0,0,0)}\MS{Check the blue text later}
\begin{equation}
    \mathcal{W}(\vx,\dot{\vx},s) = \frac{1}{2}\dot{\vx}^\tp \vM(\vx)\dot{\vx} + \varphi(\vx) + s.
    \label{storage_function_s}
\end{equation}
By construction, the storage function in~\eqref{storage_function_s} is positive definite, radially unbounded, and vanishes at the equilibrium $(\vx, \dot{\vx},s) = (\bm{0},\bm{0},0)$. To passify the closed loop dynamics, we rewrite the control law~\eqref{VSDS_new} as
\begin{equation} \label{VSDS_passive}
    \vf_{vs}(\vx) = \vPhi(\vx) + \gamma(z,s)\kappa(||\vx||)\left(\vf_{vs,o}(\vx) + \vf_{f}(\vx)\right)
\end{equation}
where
\begin{equation}
	\begin{cases}
		\gamma(z,s) = \beta(z,s) \quad z \geq 0 \\ 
		\gamma(z,s) \geq \beta(z,s) \quad \text{otherwise}
	\end{cases}.
	\label{gamma}
\end{equation}
By taking the time derivative of~\eqref{storage_function_s}, it holds that
\begin{equation}
	\begin{split}
    \dot{\mathcal{W}} &= \dot{\vx}^\tp \vM\ddot{\vx} + \nabla\varphi^\tp\dot{\vx} + \dot{s}\\
    									&= -\dot{\vx}^\tp \vD(\vx) \dot{\vx} + \dot{\vx}^\tp \vF_{ext} + \gamma(z,s) z + \dot{s}
    \end{split}
    \label{passive_storage_function_derivative}
\end{equation}
By substituting~\eqref{s_dot} in \eqref{passive_storage_function_derivative}, we obtain
\begin{equation}
\begin{split}
    \dot{\mathcal{W}} = &\dot{\vx}^\tp \vF_{ext} +(\alpha-1)\dot{\vx}^\tp \vD(\vx) \dot{\vx} +  (\gamma - \beta) z \\
    & - (\eta - \kappa(\Vert \vx \Vert))s \leq \dot{\vx}^\tp \vF_{ext},
    \end{split}
    \label{passive_storage_last_derivative}
\end{equation}
where owing to the definitions of $\alpha$, $\gamma$ and $\beta$ in \eqref{eq:conditions_alpha}, \eqref{gamma} and \eqref{beta}, respectively, we always have $s\geq0$, and in consequence the sign of the last three terms in $\dot{\mathcal{W}}$ is negative semi-definite. This results in 
$\dot{\mathcal{W}}\leq \vF_{ext}$, and we can therefore conclude the passivity of the closed-loop system with respect to the port $(\dot{\vx},\vF_{ext})$ through which the robot interacts with the external environment. 
% \YM{@Matteo: Added some explanation here, please check}
 %From the definitions of $\gamma(z,s)$ and $\beta(z,s)$, it is straightforward to verify that $\dot{\mathcal{W}} \leq \dot{\vx}^\tp \vF_{ext}$ and that the system is passive. More in details, if $z<0$ then $\gamma(z,s) \geq \beta(z,s) \geq 0$. If $z \geq 0$, the robot uses the energy in the tank to execute the non passive motion and suppresses it ($\gamma(z,s) = \beta(z,s) = 0$) if all the energy is depleted.

\subsection{Stability analysis of the passification control}\label{subsec:stability}
The storage function~\eqref{storage_function_s} is radially unbounded, positive definite, and vanishes at the equilibrium. Hence, it can be used as candidate Lyapunov function to verify the asymptotic stability of the system in the absence of external forces (i.e., $\vF_{ext} = \bm{0}$). It is worth mentioning that the passivity of the system is sufficient to ensure stability. Indeed, $\dot{\mathcal{W}}$ in~\eqref{passive_storage_last_derivative} is negative semi-definite as it vanishes for $(\dot{\vx}, s) = (\bm{0}, 0)$ irrespective of the value of $\vx$, i.e., $\vF_{ext} = \bm{0} \rightarrow \dot{\mathcal{W}} \leq 0, \forall (\vx,\bm{0},0)$. Here, we show that the system asymptotically converges to the desired goal (assumed to be $(\vx, \dot{\vx},s) = (\bm{0},\bm{0},0)$ without loss of generality). 

Let's assume that $\vF_{ext} = \bm{0}$, $\dot{\vx} = \bm{0}$, and $s=0$. This also implies that $z=0$. The closed-loop dynamics~\eqref{closed_loop_dyn} becomes
\begin{equation}
    \ddot{\vx} = \vM(\vx)^{-1}\left(\vPhi(\vx) + \gamma \kappa\left(\vf_{vs,o}(\vx) + \vf_{f}(\vx)\right) \right),
    \label{closed_loop_dyn_equilibrium}
\end{equation}
From the definition of $\gamma$ in~\eqref{gamma}, we have that $\gamma = \beta$ if $z=0$. Moreover, being $s=0$, we have from~\eqref{beta} that $\gamma = \beta = 0$ and that~\eqref{closed_loop_dyn_equilibrium} vanishes if and only if $\vM(\vx)^{-1}\vPhi(\vx) = \bm{0} \leftrightarrow \vx = \bm{0}$. The LaSalle's invariance principle~\cite{Slotine91} can be used to conclude the asymptotic stability. 

\subsection{Simulation results}
\label{sec:sim}
To verify the validity of the above theoretical concepts, we conduct a series of simulations, where we tested the developed controllers on a simulated mass. In particular, we constructed our VSDS controller with a first order DS learnt based on two different types of motions. For the angle shaped motion of the LASA HandWriting dataset~\cite{SEDS}, we compare the performance of the original VSDS approach~\eqref{eq:VSDS_org}, with the new VSDS formulation~\eqref{VSDS_passive} where the feedforward term was designed via the QP optimization~\eqref{eq:ff_qp}. On the other hand, we compare the original VSDS with the feedforward terms designed based on the velocity feedback approach~\eqref{eq:VSDS_tracking} for the curve shape.  As shown in Fig.~\ref{fig:simres}, the new formulation results in asymptotic convergence of the mass to the global attractor regardless of the method used for computing the feed-forward terms, in comparison to a very low steady state error in the original formulation. Figures~\ref{fig:angle_tank} and~\ref{fig:tank_curve} show the tank state $s$, where it could be noted that the tank is slowly depleted in the beginning of the motion, followed by a more rapid convergence when the linear dynamics become more dominant close to the equilibrium. \\
In the second simulation study, we compare the performance of our VSDS QP approach to the controller proposed in \cite{kronander2015passive}. In particular, we showcase the benefit of the encoded spring attraction in following a nominal path and velocity profile. The controller in \cite{kronander2015passive} guarantees a passive interaction behavior, and can be also used to follow any first-order DS, with a control law formulated as  $\vF=\vD_{PI}(\vx)(\vf_g(\vx)-\dot{\vx})$. The damping matrix $\vD_{PI}$ is designed to have the first eigenvector aligned with the direction of motion, while the remaining eigenvectors are perpendicular to each other.
We construct a first-order DS using the demonstrations of a W-shape, and compare the performance of both controllers when subjected to perturbations. We tune both controllers to have a good tracking performance of the desired path and velocity in the free motion case, as well as a similar disturbance response (i.e., maximum deviation from path when subjected to a disturbance). To simulate the disturbance, we apply a constant force of $25\,$N for $2\,$s, pointing in the direction perpendicular to the current motion direction. As shown in Fig.~\ref{fig:w_spatial_comp}, the mass driven by the controller in~\cite{kronander2015passive} does not return back to the reference path after the disturbance vanishes, and instead follows a shortcut path to the goal. This is also reflected at the velocity level, as shown in Fig. \ref{fig:w_vel_x_comp} and \ref{fig:w_vel_y_comp}. These problems are remedied in the proposed QP controller, where the mass is pushed back to the reference, and also correctly tracks the desired velocity after recovering from the applied disturbance. 

\section{Results}\label{sec:experiments}
\begin{figure*}[t]
\centering
	
	\includegraphics[width=0.31\textwidth]{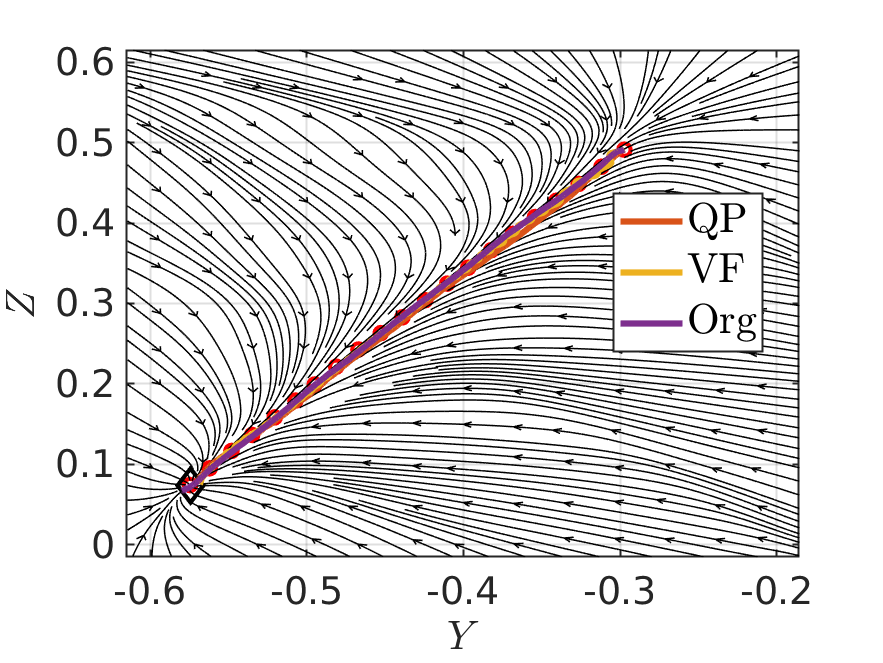}
	\includegraphics[width=0.31\textwidth]{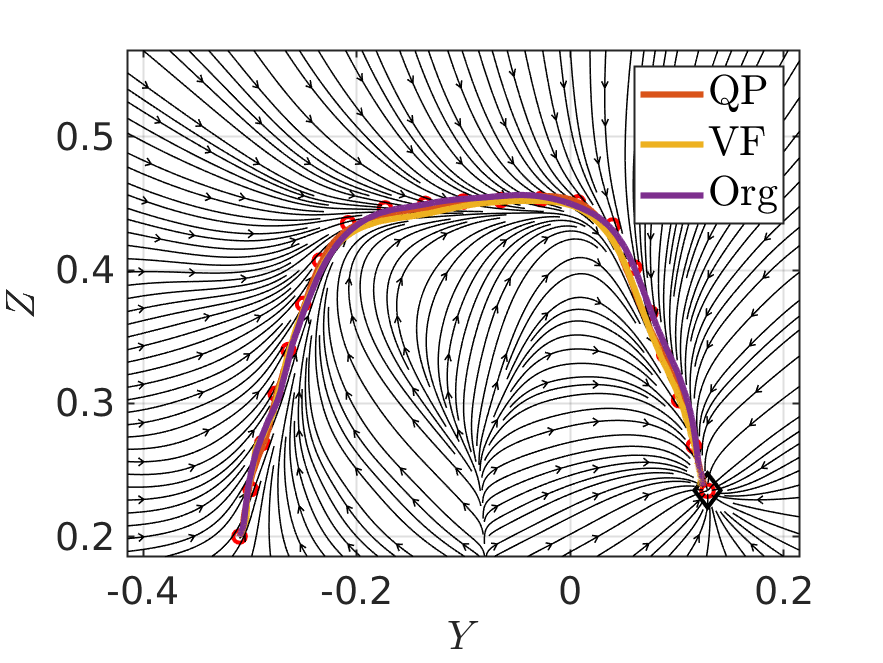}
	\includegraphics[width=0.31\textwidth]{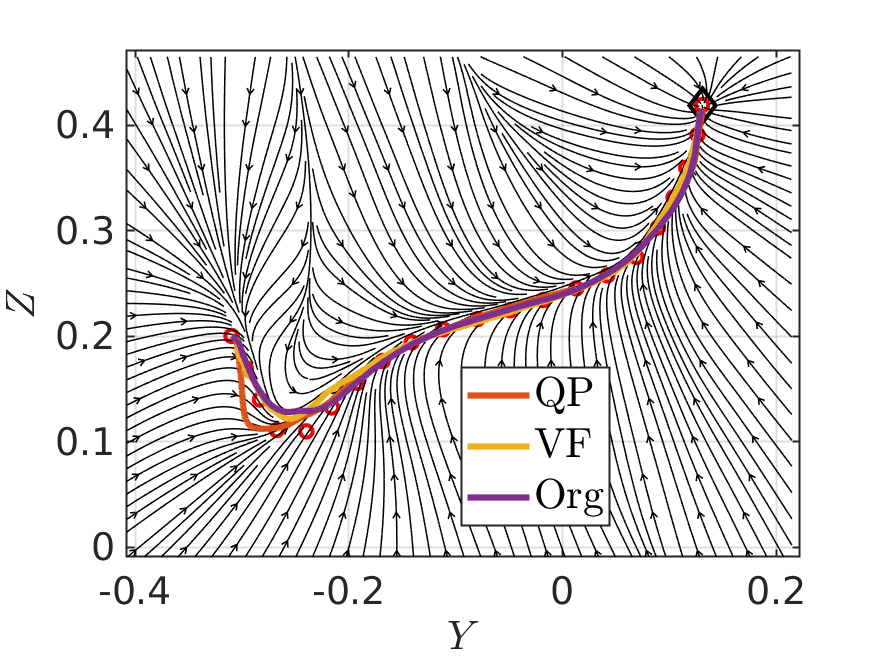}
 \\
 \includegraphics[width=0.31\textwidth]{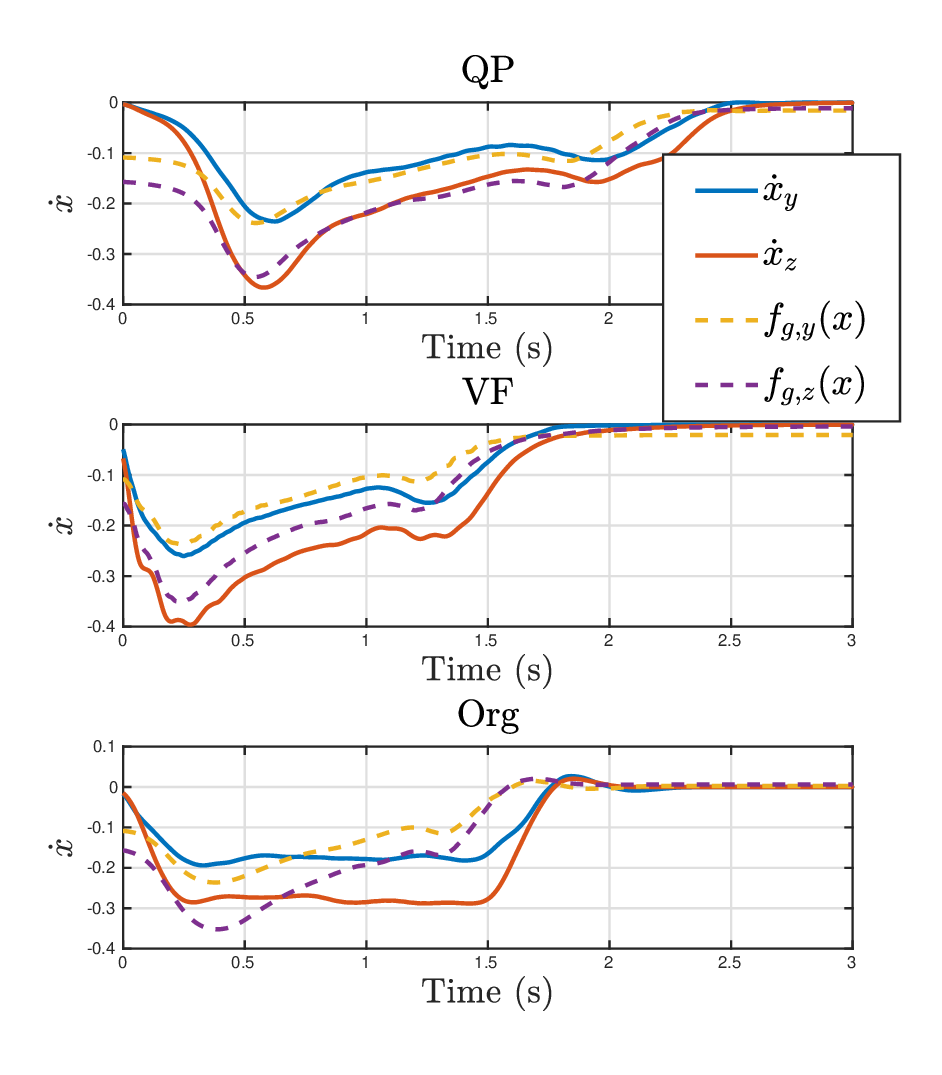}
 \includegraphics[width=0.31\textwidth]{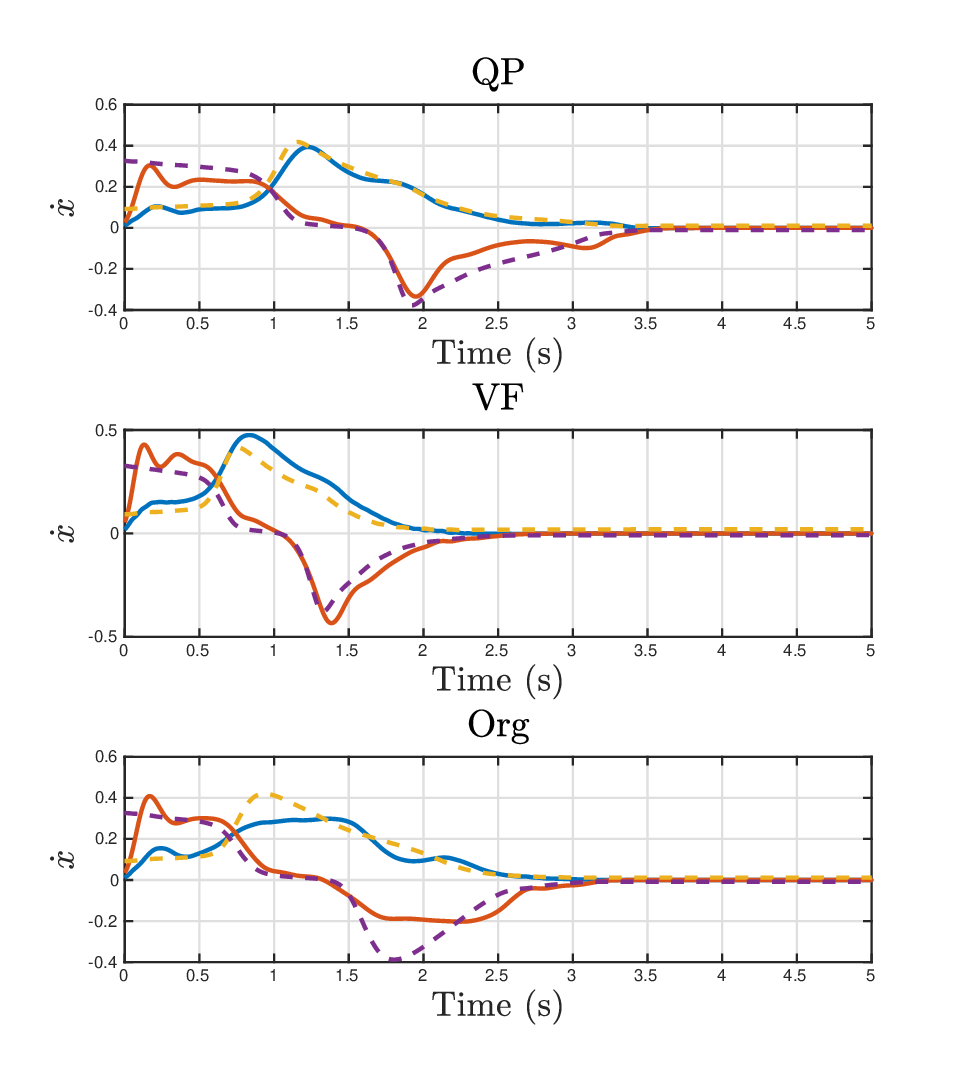}	
  \includegraphics[width=0.31\textwidth]{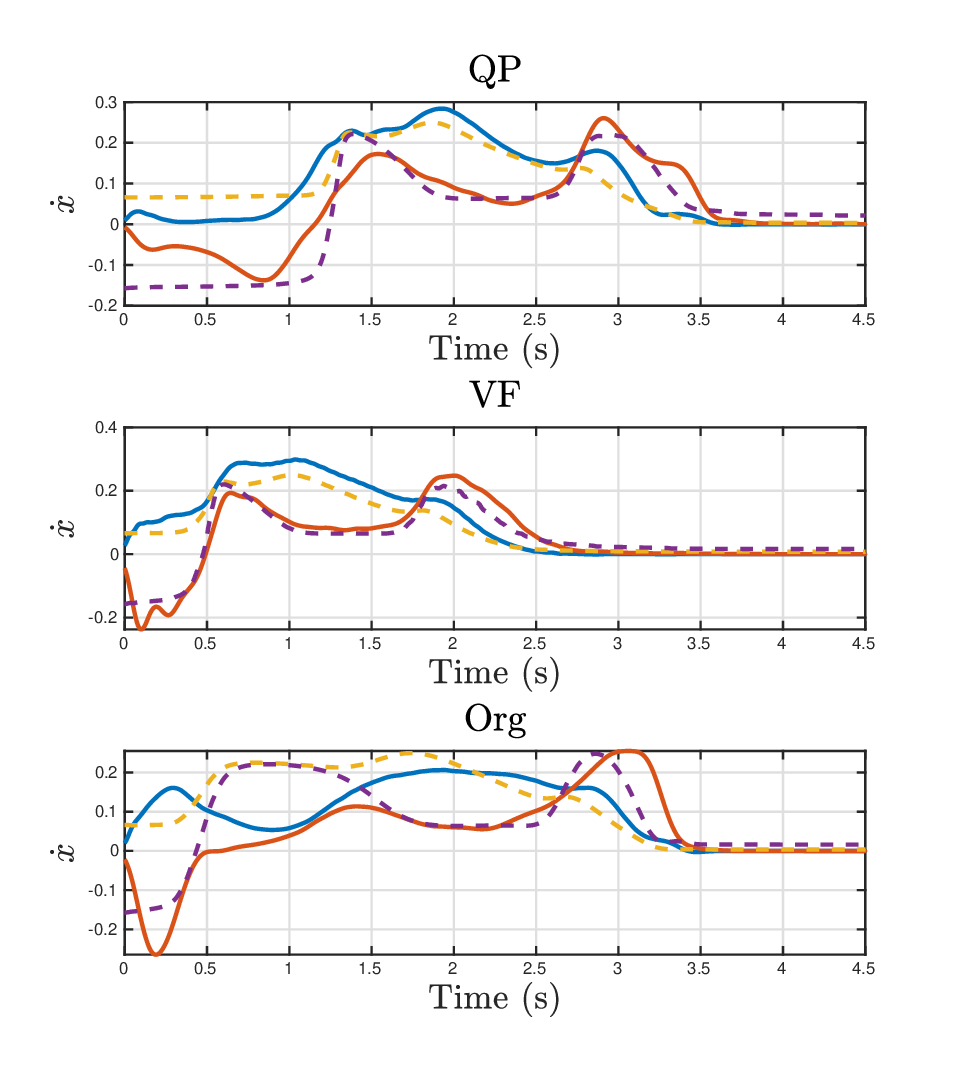}
  
 	\caption{Experimental results for the controllers comparisons for motion execution. The first row shows the spatial position of the executed motions for the constant stiffness case, with the streamlines of the VSDS dynamics in the background. The lower row shows the corresponding velocity profile with solid lines as the actual velocity and dotted as the desired, for the $y$- and $z$- directions. }
	\label{fig:MotExec}
\end{figure*}

\begin{figure}[!t]
\centering
	\subfigure [RMS Velocity Error]{
	\includegraphics[width=0.225\textwidth]{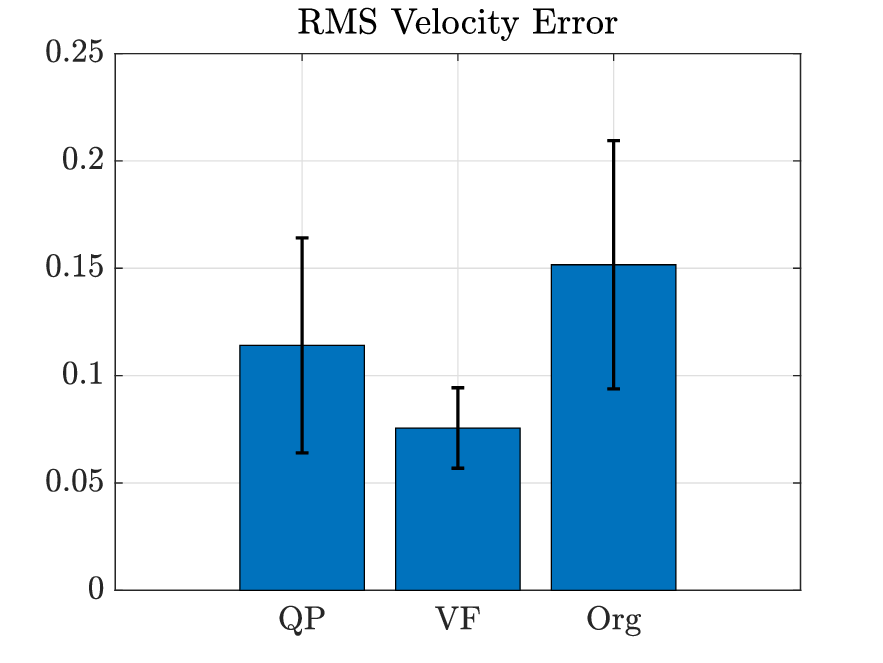}
		\label{fig:rms}
	}
	\subfigure [Tank state]{
	\includegraphics[width=0.225\textwidth]{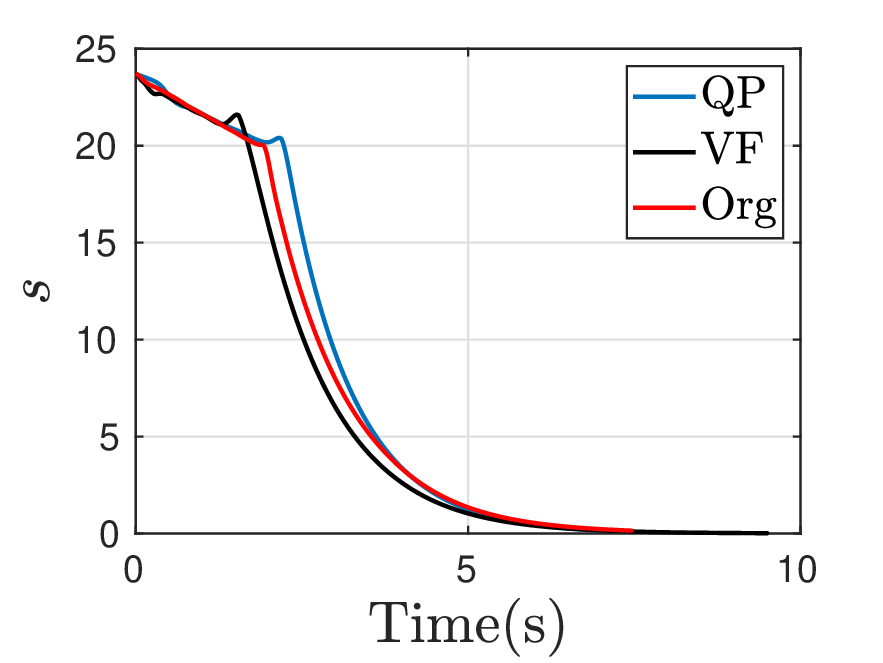}
		\label{fig:tank}
	}
 \caption{Left: Mean RMS velocity error over all executed motions for each controller. Right: comparison of the tank states from the three controllers, for the same executed motion type and stiffness condition.}
	\label{fig:tankrobandrms}
\end{figure}

\begin{figure*}[!t]
\centering
    \subfigure [Initial Configuration]{
	\includegraphics[width=0.225\textwidth]{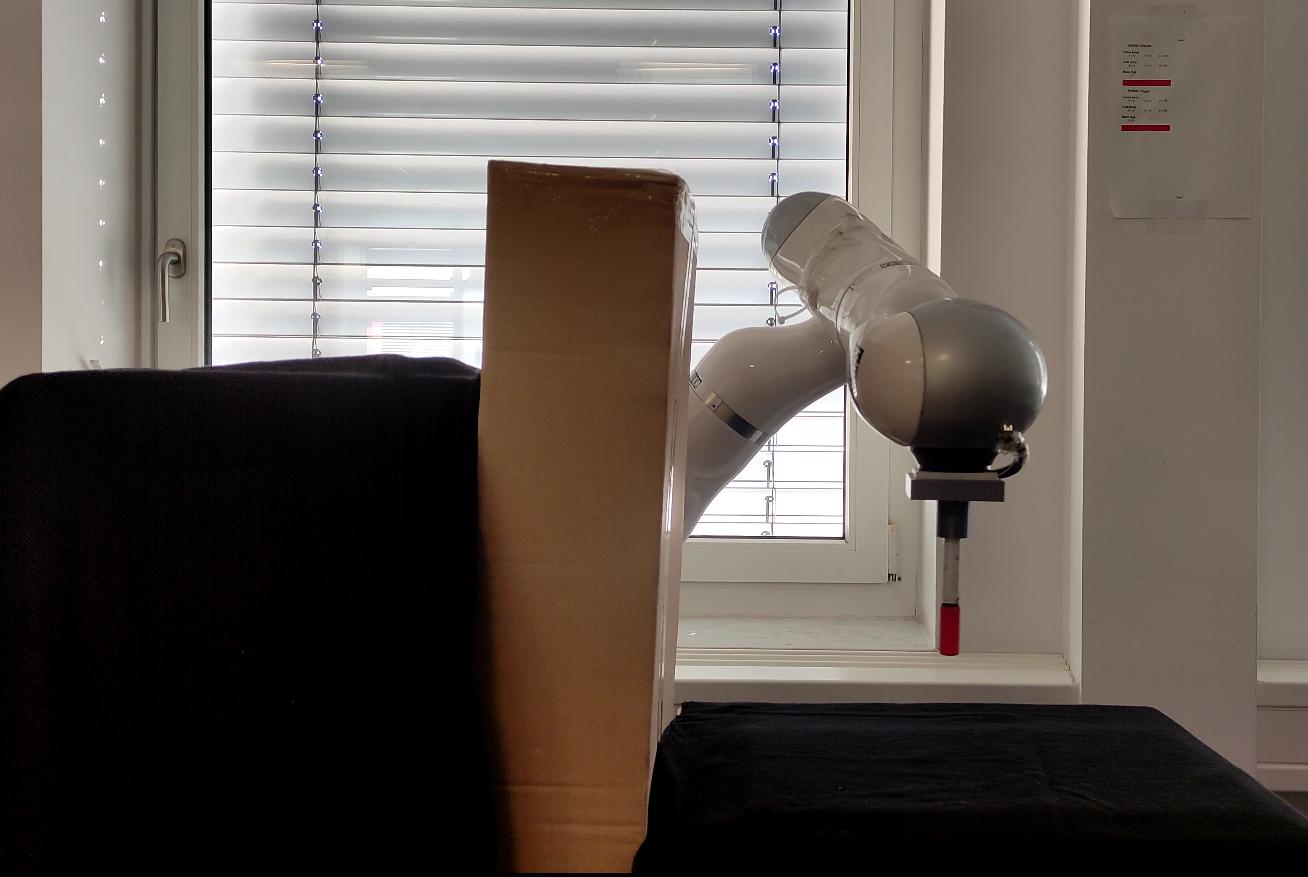}
		\label{fig:crash_set_init}
	}
    \subfigure [Final Config.-Original VSDS]{
	\includegraphics[width=0.225\textwidth]{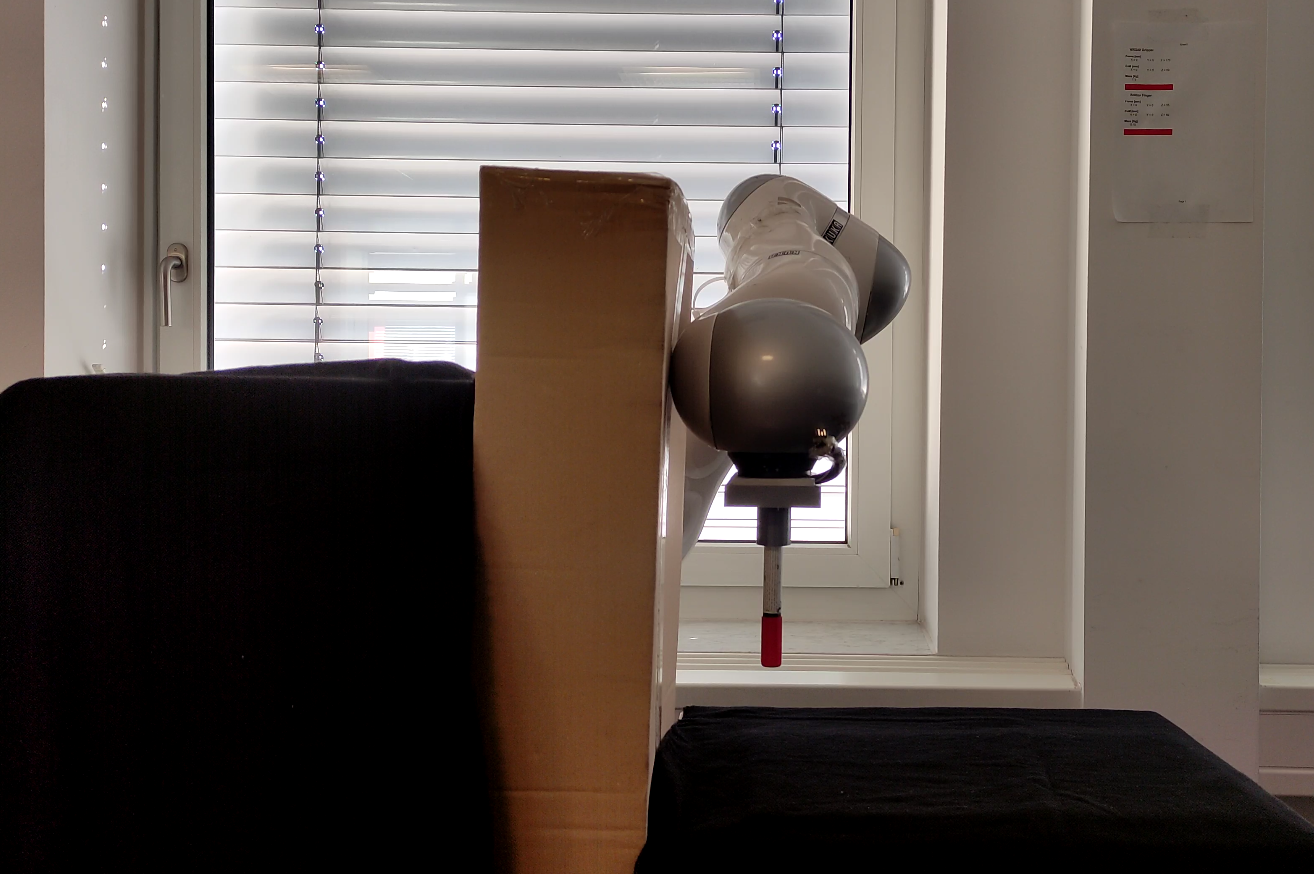}
		\label{fig:crash_set_org}
	}
    \subfigure [Final Config.-QP ]{
	\includegraphics[width=0.225\textwidth]{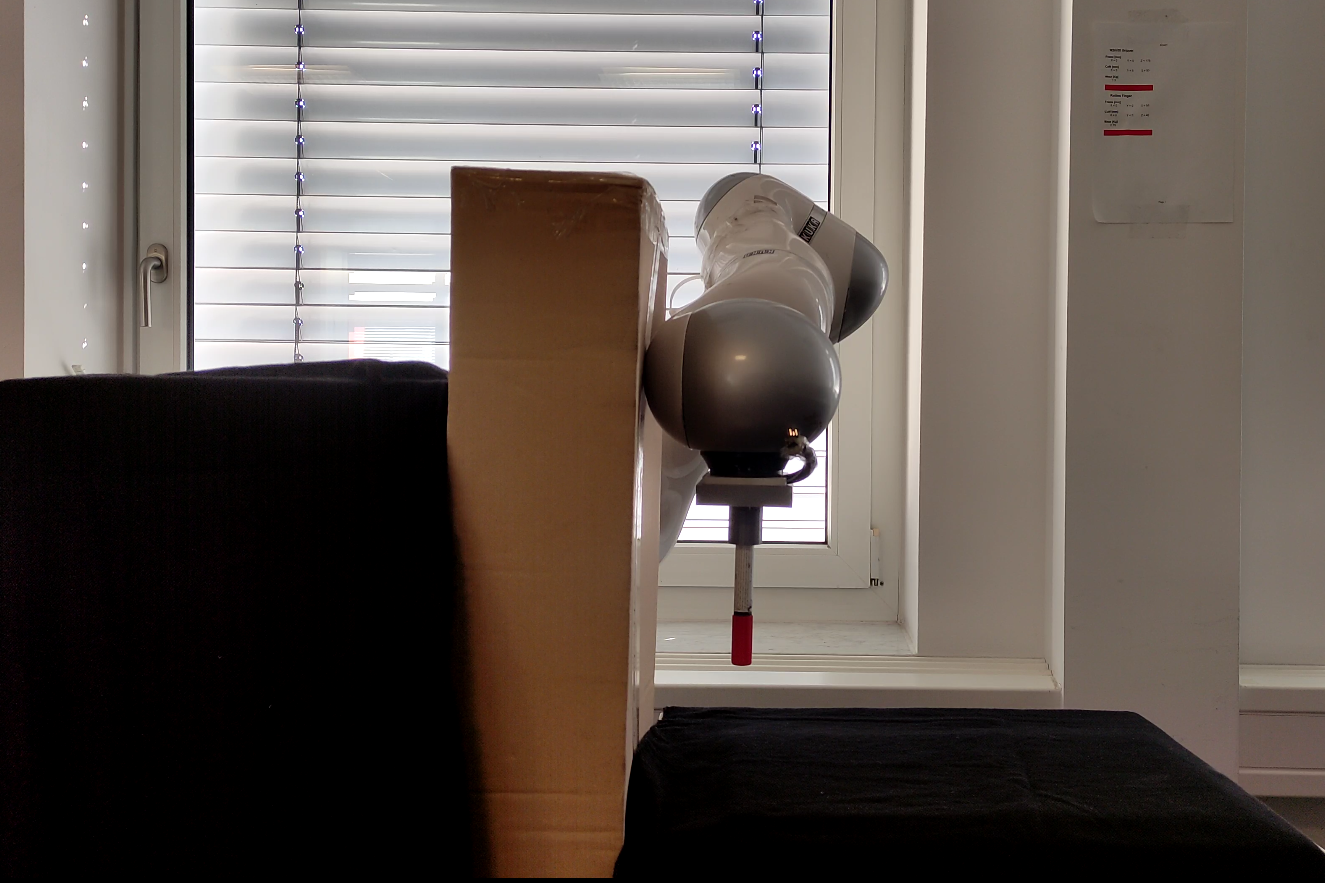}
		\label{fig:crash_set_qp}
	}
    \subfigure [Final Config.-VF ]{
	\includegraphics[width=0.225\textwidth]{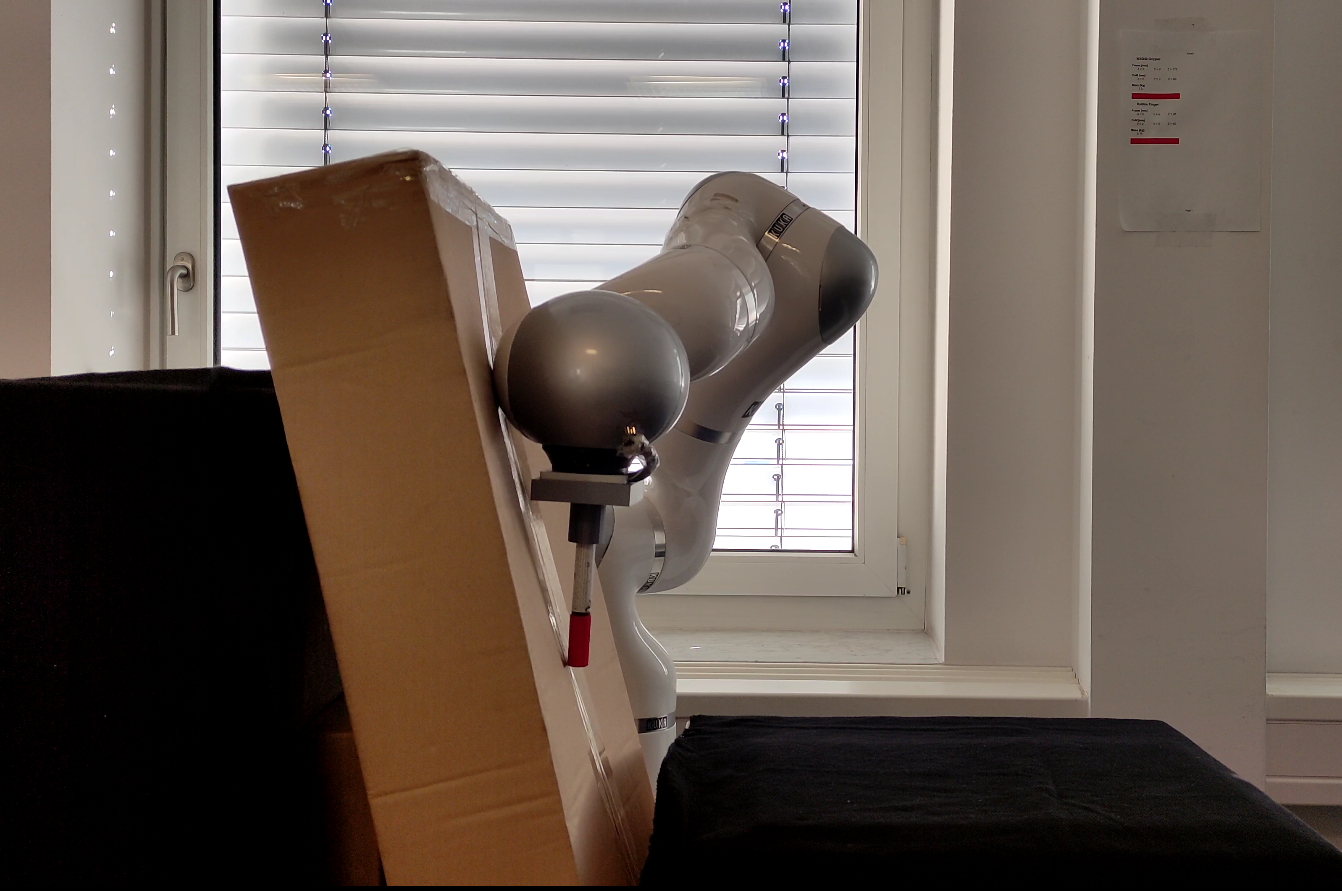}
		\label{fig:fig:crash_set_vf}
	}
	\subfigure [Position over time]{
	\includegraphics[width=0.3\textwidth]{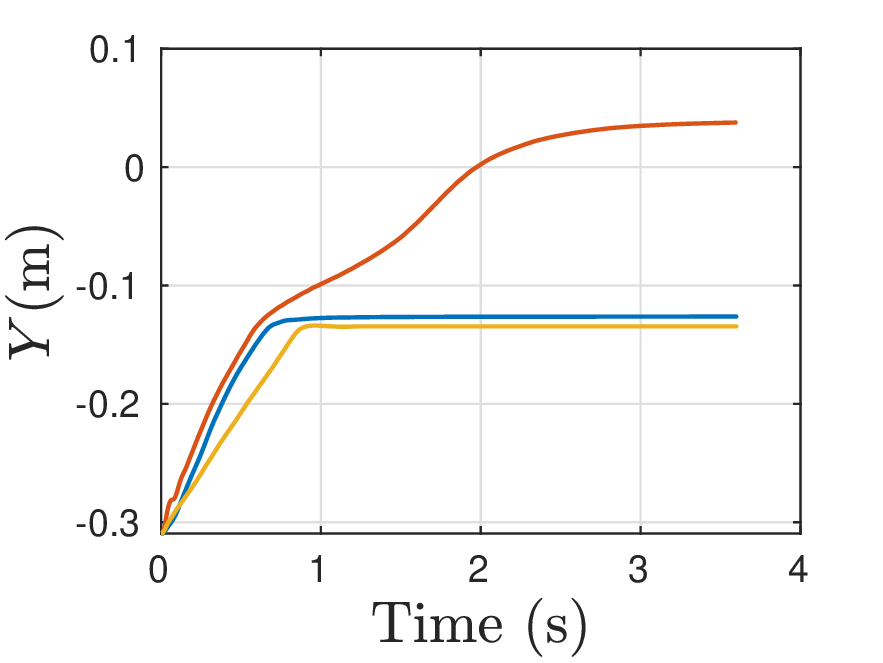}
		\label{fig:coll_pos}
	}
	\subfigure [External Force]{
	\includegraphics[width=0.3\textwidth]{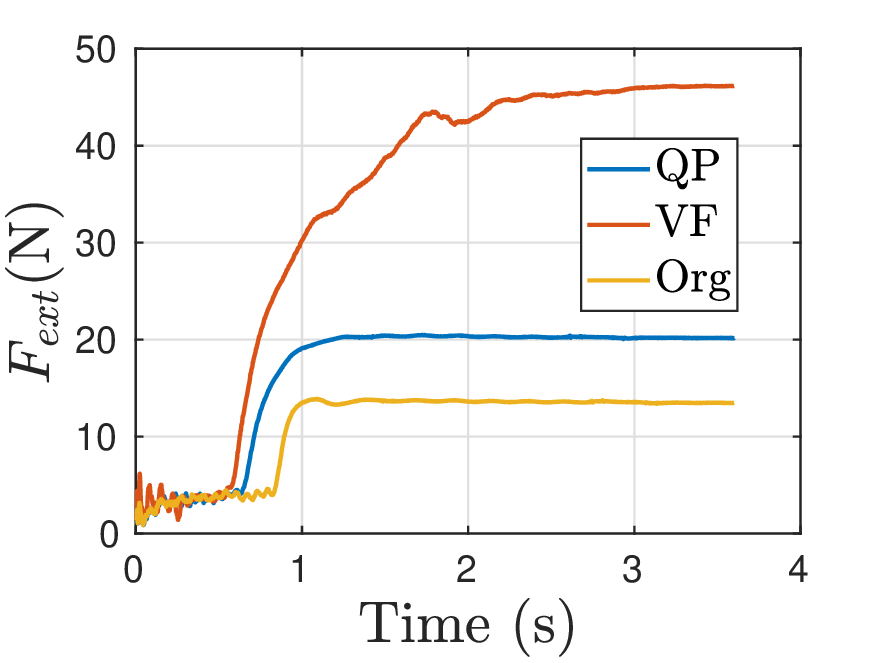}
		\label{fig:coll_force}
	}
 \caption{Results from the Collision experiments. Top row shows a snapshot from the setup at the beginning of the experiment, as well as a snapshot from the final configuration for each of the compared approaches. The bottom row shows the position and the norm of the external force for the three approaches.  }
	\label{fig:collisions}
\end{figure*}
In this section, we conduct a series of experiments in order to validate our approach in terms of accurate motion execution, safety and a physical interaction tasks. The validation is performed on a 7-DOF Kuka LWR, controlled via a Desktop PC with a Core i7. The robot is commanded via the Fast Research interface library, in Cartesian Impedance control mode where the computed control law $\vF$ is sent to the robot as a feed-forward force. Unless otherwise stated, in the following evaluations we used $N=20$ linear DS. For the potential $\vPhi(\vx)$, we set $\zeta=0.006$, $\tau_{min}=1$, while $k_o$ needed slight tuning depending on the VSDS approach used, but was typically set in the range of $0.3$ to $0.5$. Similarly, the damping gains $\vD_i$ used in \eqref{eq:VSDS_tracking} for the Velocity Feedback Method were set in the range $200-300$ \si{N}{/m}$^2$, depending on the motion type to be executed. For the tank, we used $s_0=30$ and $\eta=1.05$. Based on practical experience, we set the values of $\underline{\vF}$ in the optimization problem \eqref{eq:QP} to $10$\si{N}, which represents a high enough initial force to allow a Kuka LWR to move. 
\subsection{Motion Execution} 
In the first part of the validation, we test the ability of our controller to execute motions following a desired path and a reference velocity profile, while also asymptotically converging to the global attractor.
% We compare our original VSDS \cite{}, to the two approaches presented in this paper, namely the Velocity Feedback and the QP-based optimization (referred to VF and QP from hereon) methods for computing the feed forward forces of equation \eqref{VSDS_new}. 
We use \eqref{VSDS_passive} to compute the VSDS force component $\vf_{vs}(\vx)$, subsequently used for commanding $\vF$ as in \eqref{F_new}. We compare with $\vf_f(\vx)$ designed based on Velocity Feedback (VF) (eqn. \eqref{eq:VSDS_tracking}), the optimization-based design (QP) (eqn. \eqref{eq:ff_qp}) and the original VSDS approach (Org) with $\vf_f(\vx)=\mathbf{0}$. 

As stated earlier, the specification of the stiffness should be independent from the velocity profile of the robot. Therefore, we compare the motion execution for two stiffness profiles: a constant stiffness with $\vK_d=diag(1200,1500)$ and a state-varying stiffness with diagonal elements set to $950+150$sin$(15\vx^1+0.8)$ and $1200+200$sin$(15\vx^1+0.8)$. Finally, we test three motions from the LASA Handwriting Data Set with increasing levels of complexity~\cite{SEDS}: a straight line, a trapezoidal motion and a Khamesh Shape. The motion data was first appropriately scaled and shifted to make it feasible for robot execution, then we used the approach from \cite{pmlr-v87-figueroa18a} to learn the first-order asymptotically stable DS $\vf_g(\vx)$, required for computing VSDS. To conclude, we conducted a total of 3$\times$3$\times$2 experiments on the robot. \\
The results of the experiments are shown in Fig. \ref{fig:MotExec} and \ref{fig:tankrobandrms}. The first row in Fig. \ref{fig:MotExec} shows the spatial motion in the constant stiffness case\footnote{The spatial motion in the varying stiffness case looks exactly the same, and therefore we omit it for brevity} overlaid on the streamlines of the original VSDS. The second row shows the actual robot velocity compared to the desired one computed by $\vf_g(\vx)$ for the constant stiffness case. In Fig. \ref{fig:rms}, we show the mean and standard deviation of the Root Mean Square (RMS) velocity error ($\dot{\ve}$)  for the three controllers over all the executed motions (6 for each controller). Finally, Fig. \ref{fig:tank} shows an example of the tank state from the same condition for the three control formulations. \\
  For the original VSDS, the actual robot velocity is clearly different from the desired velocity profile $\vf_g(\vx)$. This gets resolved by adding the feed-forward term $\vf_f(\vx)$, which improves the tracking accuracy of the reference velocity, and where the VF approach yields best tracking results, reflected by the lowest mean for the RMS error. On the other hand, all three controllers are able to guide the robot to the global attractor, with the tank rapidly converging to zero close to the equilibrium, which is also consistent with the simulation results of IV.D. 
  
  % Note however, that in contrast to the derived stability proof of (), some energy still remains in the tank. This is due to a very small steady state error (typically less than \SI{3}{cm}) which is possibly caused by friction and umodelled dynamics. This can be confirmed by noting that simulation results with the QP controller for the same experiment, the tank energy converges to 0. 

\begin{figure*}[t]
\centering
	\subfigure [Robot Setup]{
	\includegraphics[width=0.13\textwidth,height=0.17\textwidth]{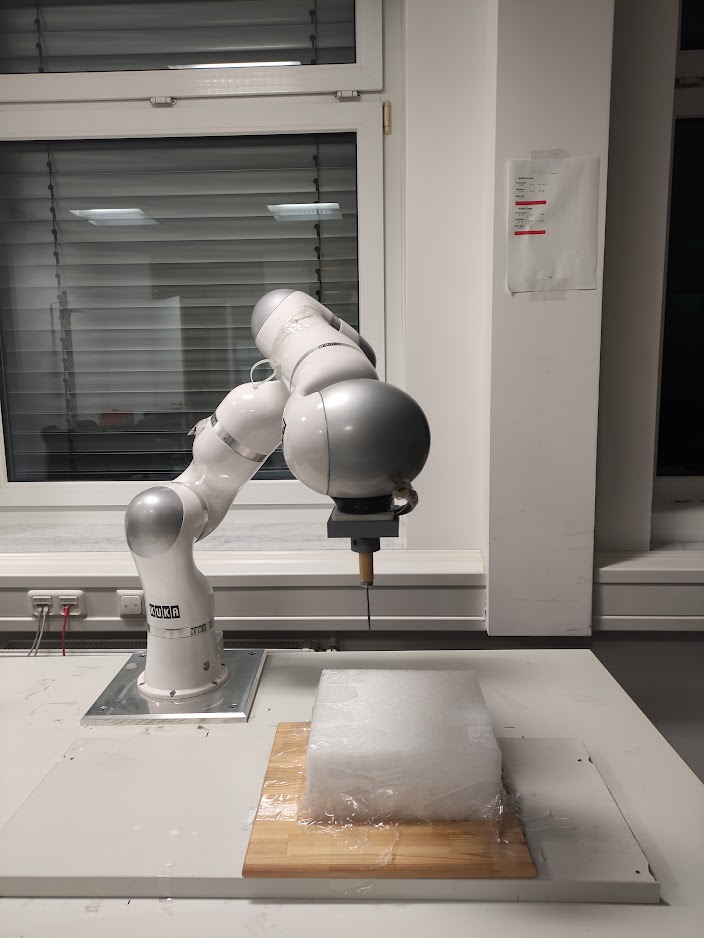}
		\label{fig:setup_drill}
	}
	\subfigure [Spatial Position ]{
	\includegraphics[width=0.25\textwidth]{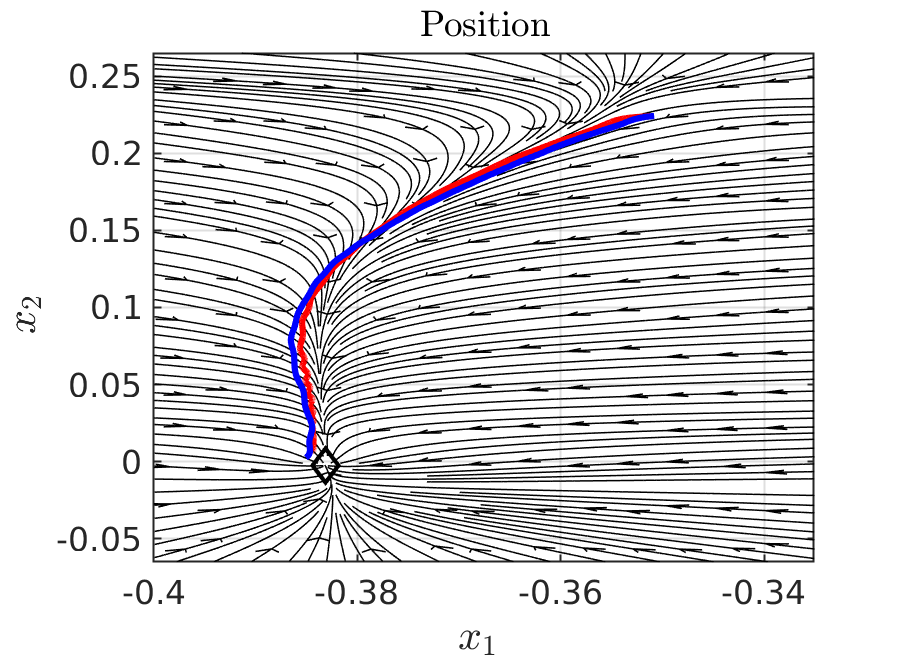}
		\label{fig:Pos_Drill}
	}
 	\subfigure [Desired vs actual velocity]{
	\includegraphics[width=0.25\textwidth]{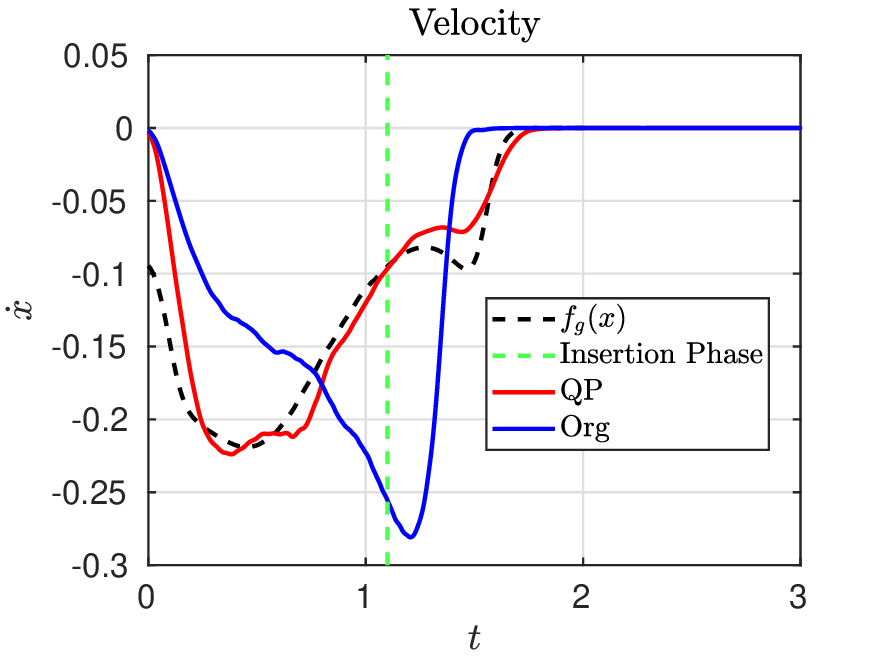}
		\label{fig:Vel_Drill}
	}
	\subfigure [External Force norm]{
	\includegraphics[width=0.25\textwidth]{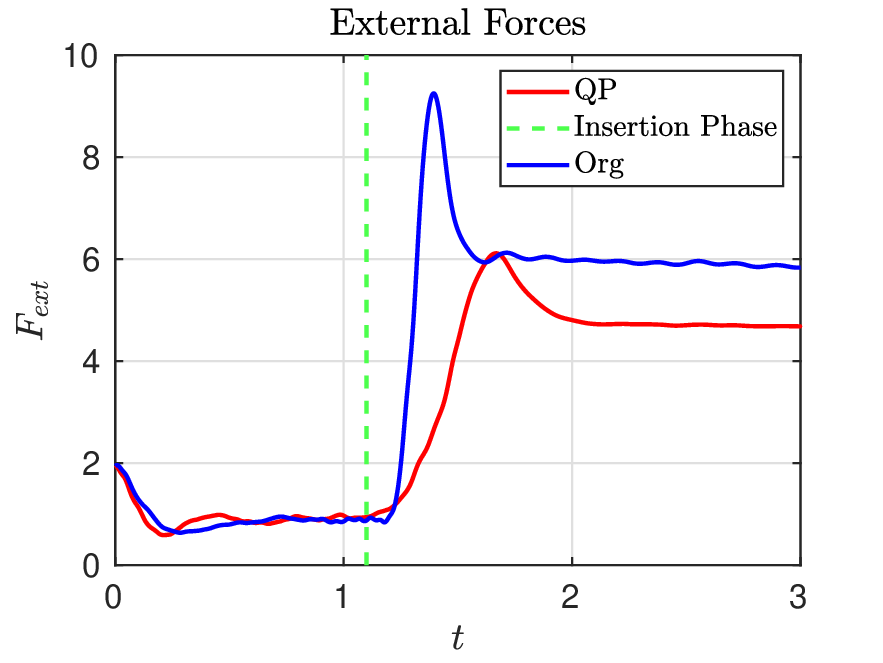}
		\label{fig:force_drill}
	}\\

 \caption{Experimental results of our drilling-like task, with the first figure showing the experimental setup of the robot, while the last three figures show the results of the task execution for the QP (red) and the Org.(blue) VSDS approaches.   } 
	\label{fig:drill}
\end{figure*}

\begin{figure*}[t]
\centering
 \subfigure [Snapshot during HRI]{
	\includegraphics[width=0.165\textwidth]{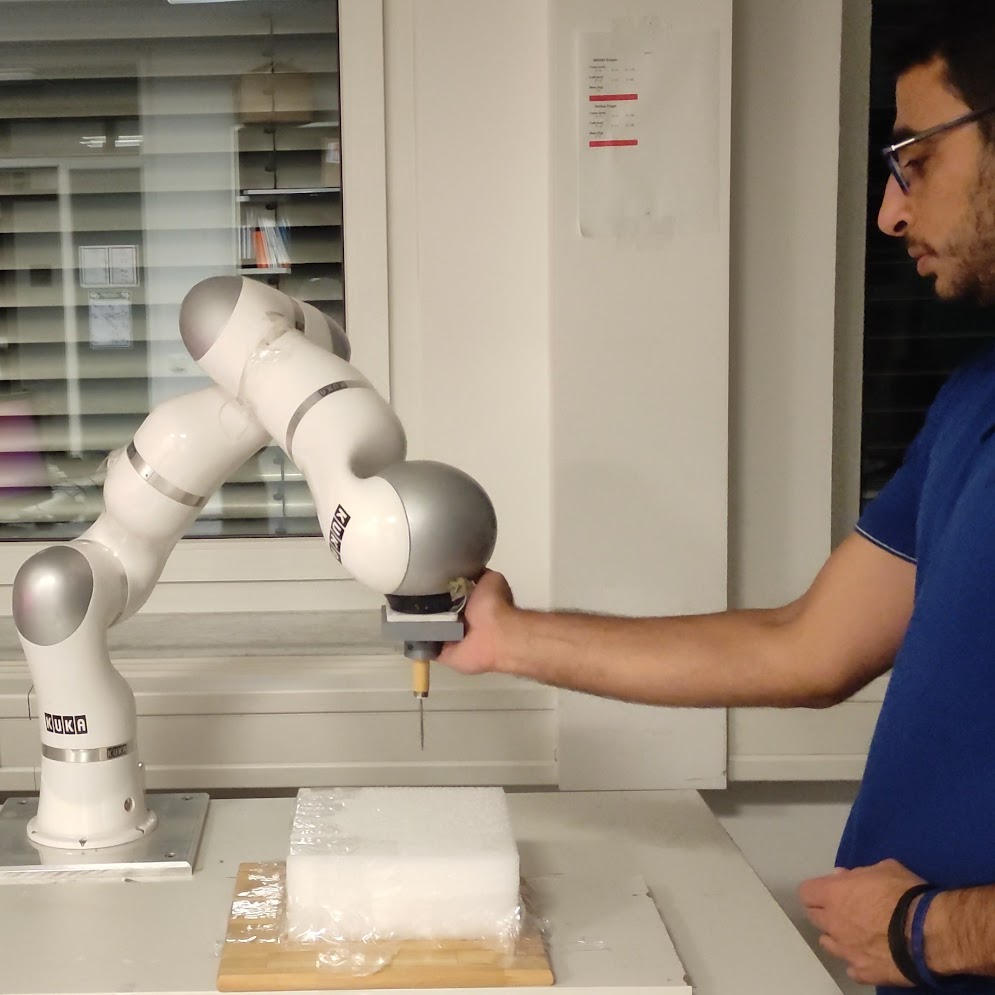}
		\label{fig:setup_drill_HRI}
	}
 \subfigure [Spatial Position during HRI]{
	\includegraphics[width=0.25\textwidth]{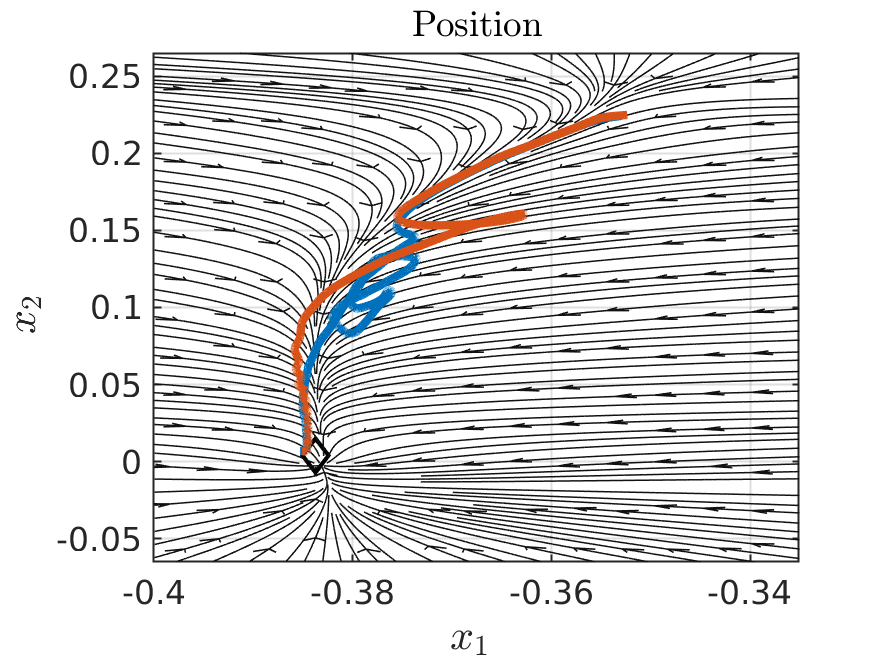}
		\label{fig:Pos_Drill_HRI}
	}
 	\subfigure [Velocity during HRI]{
	\includegraphics[width=0.25\textwidth]{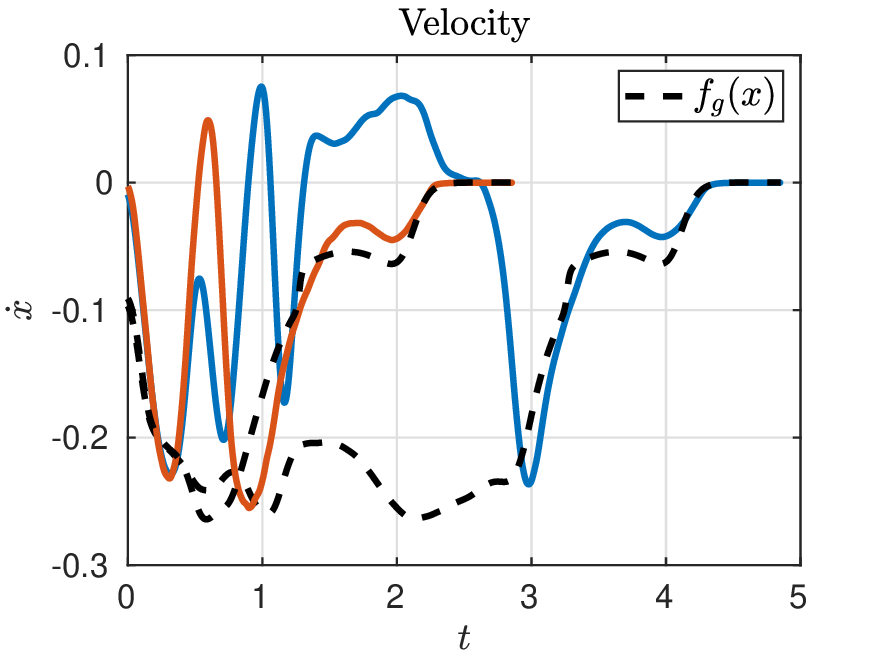}
		\label{fig:Vel_Drill_HRI}
	}
	\subfigure [External Force norm during HRI]{
	\includegraphics[width=0.25\textwidth]{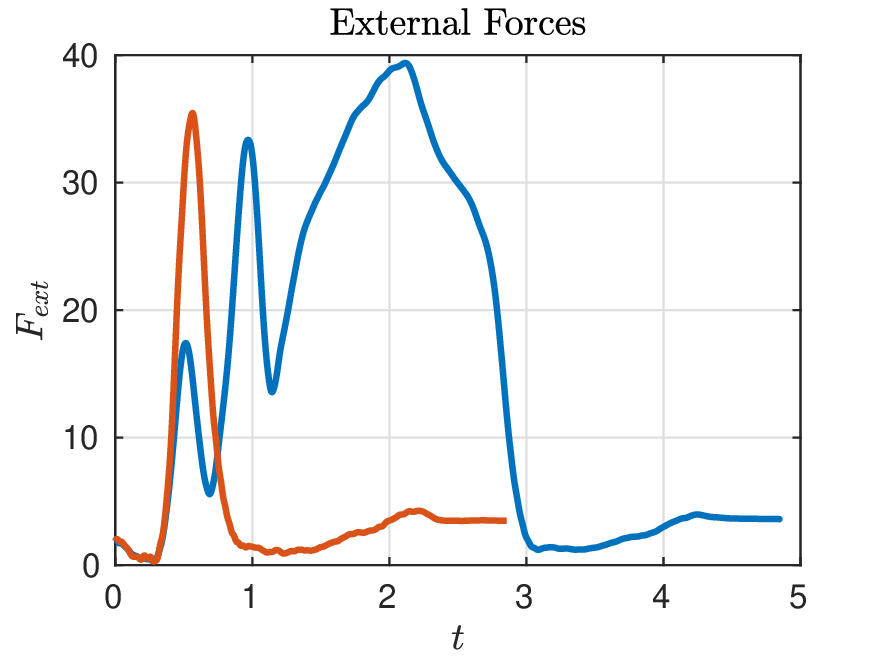}
		\label{fig:force_drill_HRI}
	}
 \caption{ Experimental Results of the drilling task for the QP approach where the robot is subjected to perturbations applied by a human. The first figure shows a snapshot where a human physically interacts with the robot, while the remaining figures show the spatial position, velocity and external forces from two different HRI experiments, depicted in orange and blue. 
  } 
	\label{fig:drillHRI}
\end{figure*}

\subsection{Safety} 
In the second experiment, we validate the safety of the presented approaches in unexpected collisions, by placing a carton box in the path of the planned robot motion, as shown in Fig. \ref{fig:crash_set_init}. We learned a first-order DS based on a straight line minimum jerk trajectory in the $y$- direction, and use it to construct our VSDS, where we also compare the same three VSDS variations from the previous subsection. The results of this experiment are shown in Fig. \ref{fig:collisions}. Clearly, as can be also shown in the video, the interaction is safe for the original VSDS and the QP cases, highlighted by the relatively low external force (Fig. \ref{fig:coll_force}), with a slightly higher force for the QP case. Note also the fact that the robot does not "push" the carton box, which can be verified by the steady state robot position (Fig. \ref{fig:coll_pos}) for these two cases is right at the collision point with the box (Fig. \ref{fig:crash_set_org}, \ref{fig:crash_set_qp}). On the contrary, the robot keeps on moving against the box (Fig. \ref{fig:fig:crash_set_vf}) for the VF case, which results in a much higher collision force. This effect can be mainly attributed to the feed-forward term in equation \eqref{eq:VSDS_tracking}, where increasing damping gains to a certain extent improves the velocity tracking, however at the expense of a higher steady state control force, which eventually increases the collision force. We would like to note however that while the magnitude  steady state external force in the VF case was close to the external force we noticed in our previous work \cite{chen2021closed} in the case of a time-indexed trajectory, the VF controller is still safer in the sense the control force does not increase over time, which typically results in aggressive robot motions once the obstacle blocking the robot is removed.   

\subsection{Interaction Task}
In this experiment, we tested our approach in a simplified drilling-like task which requires the robot to penetrate a foam surface with a needle-like tool mounted on its end-effector ( Fig. \ref{fig:setup_drill}). For such a task, the robot starts from an initial position above the surface, approaches the drilling point with an arbitrary velocity and ideally maintains a constant low velocity during the insertion phase and therefore, following a specific velocity profile would be desirable. A human provides demonstrations to the robot in gravity compensation mode, while recording the end-effector position, and obtaining the velocities via finite differences, which serve as training data to learn a first-order DS with SEDS  \cite{SEDS}. This is then used to construct our VSDS, where we use a state-varying stiffness profile that starts with a constant stiffness of \SI{1000}{N/m}, and increases smoothly to $\SI{1800}{N/m}$ with a minimum jerk trajectory slightly before approaching the insertion location to compensate for the physical interaction. We compare the performance of our original VSDS approach, with the QP approach for the design of the $\vf_f(\vx)$. As can be shown in Fig. \ref{fig:drill} and in the attached video, the task can be completed with both approaches. Note  for the QP approach, the actual robot velocity follows well the desired velocity $\vf_g(\vx)$ (Fig. \ref{fig:Vel_Drill})\footnote{For clarity, we show only the velocity results in the main direction of motion which is the $z-$ axis. }, which also reflects the human strategy used during the demonstrations to maintain a constant low velocity during the insertion phase. This results in a smooth task execution, as compared to the original VSDS approach, where the robot has a velocity profile that correlates with the stiffness, increasing during the insertion phase. This leads to a larger impact and in a consequence a higher overshoot in the external force sensed at the robot end-effector can be observed, as compared to the QP approach (Fig. \ref{fig:force_drill}). 
In the second set of experiments (Fig. \ref{fig:drillHRI}), we compare the robustness of our QP control approach to perturbations, applied by a human physically interacting with the robot during task execution. As shown in the video, the robot reacts in a safe and compliant manner to the applied disturbances, and is able to resume smoothly the task execution, while still maintaining the desired velocity profile during the insertion phase (Fig. \ref{fig:Vel_Drill_HRI}).   
\section{Discussion and Conclusion}\label{sec:conclusion}
In this work, we advanced the potential of our VSDS controller in two main directions. First, we guarantee the stability of our controller by exploiting energy tanks to enforce passivity, thereby ensuring stable interactions with passive environments. We further extended our proof to include asymptotic stability with respect to a global attractor in free motion. Our second goal was to make our controller more suited for tasks that require trajectory tracking, while still benefiting from the inherit safety properties of the closed-loop configuration, as shown in our collision experiments. To achieve these goals, we proposed two formulations based on velocity error feedback and QP optimization of the feed-forward terms. The proposed formulations were validated in simulations and real robot experiments. The stability feature was highlighted in the simulation results (Fig. \ref{fig:sim_norm_curve} and \ref{fig:angle_norm}), where the proposed VSDS yielded exact convergence to the goal attractor, compared to a small steady state error for the original VSDS. On the other hand, the trajectory tracking capability was highlighted in Fig. \ref{fig:rms}  where the new VSDS formulation yielded lower velocity tracking errors compared to the original VSDS. The velocity tracking capability also proved to be beneficial in the chosen interaction task, resulting in a safer task execution, as shown by the lower interaction forces in  Fig.~\ref{fig:force_drill}. 

Generally, the QP approach seems to be safer in terms of external collisions, however, with respect to tracking, the approach resulted in higher velocity errors. Clearly, the choice between the QP and VF approaches is application dependent, and represents a safety/performance trade-off.

Since the QP approach is based on a simulated model, unmodelled robot dynamics such as friction seem to affect the velocity tracking performance and the steady state convergence, and therefore proper identification and compensation of these terms would be one way to improve performance. Another potential solution would be to obtain the training data for the QP optimization problem based on an actual robot execution of the open-loop integrated trajectory, instead of a simulated model. 
It is worth mentioning also that our QP optimization shares some similarities from regression based approaches for stiffness estimation \cite{Rozo,ABUDAKKA2018156}. Similar to our case, these works also assume a second-order model to fit the observed robot dynamics. We also exploit a second order model, however the goal for these approaches is to extract the stiffness profiles used during demonstrations via regression, while in our case, we assume the stiffness is already provided by the user, and instead aim to extract the spring force driving the simulated robot.\\
In the future, we aim to extend our VSDS formulation to include orientation tasks. In particular, we will consider the use of DS based on unit Quaternions to represent a motion plan and subsequently derive our VSDS controller. 
To this end, we will explore Riemannian manfolds \cite{Riem2,matteoriem} and their associated operations such as exponetional maps and parallel transports in order to find the right formulation for sampling the via points, constructing, and combining the linear springs around each local equilibrium.   

\section*{Acknowledgement}
This work has been partially supported by the European Union under NextGenerationEU project iNEST (ECS 00000043).
% \textcolor{red}{@Youssef: put a couple of simulations to show the stability proof works.}

% . Ideally, in an open-loop trajectory tracking problem, the velocity of the robot should be similar to that of the desired time-indexed trajectory $\vx_d(t)$, independently from the values of the stiffness and damping \footnote{With the assumption of a stiffness high enough to overcome robot friction, and properly tuned damping.}. These impedance parameters should on the other hand mainly affect the robot behavior in physical interaction, in the sense of how it reacts to perturbations or allows deviation from the desired trajectory.

% First, the robot only practically converges to the global attractor or very close to it, which is also partially dependent on parameter tuning. In other words, there is no theoretical guarantees that the robot is asymptotically stable with respect to the global attractor, which is one of the main features of first order DS. The second problem, 

%  First, we show how to modify our controller in order to ensure the 

%\section*{Acknowledgements}

\bibliographystyle{IEEEtran}
\bibliography{bibliography.bib}

\end{document}